\documentclass{article}

\usepackage{microtype}
\usepackage{graphicx}
\usepackage{subcaption}
\usepackage{booktabs} 

\usepackage{hyperref}

\usepackage[preprint]{icml2026}

\usepackage{amsmath}
\usepackage{amssymb}
\usepackage{mathtools}
\usepackage{amsthm}

\usepackage{multirow}
\usepackage{array}
\usepackage{makecell}
\usepackage{colortbl}
\usepackage[dvipsnames]{xcolor}
\usepackage{pifont}
\usepackage[most]{tcolorbox}
\newcommand{\Sref}[1]{\hyperref[#1]{\S\ref{#1}}}

\definecolor{kclred}{HTML}{E12726}       
\definecolor{kclbg}{HTML}{FDF5F5}       

\definecolor{takeaway}{HTML}{8B4513}

\newtcolorbox{titlecard}{
  enhanced,
  frame hidden,           
  arc=10pt,               
  colback=kclbg,       
  left=0.5cm,
  right=0.5cm,
  top=0.5cm,
  bottom=0.5cm,
  before skip=0pt,
  after skip=0.5cm,
  overlay={\node[anchor=south east, at=(frame.south east), 
    xshift=-0.1cm, yshift=0.1cm] 
    {
    \includegraphics[height=0.65cm]{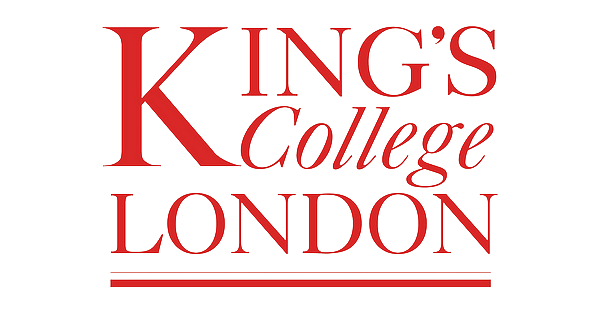}
    \hspace{-0.1cm}  
    \includegraphics[height=0.6cm]{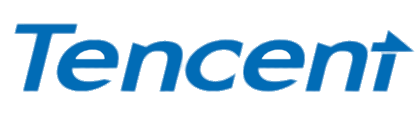}
    \hspace{-0.1cm}  
    \includegraphics[height=0.65cm]{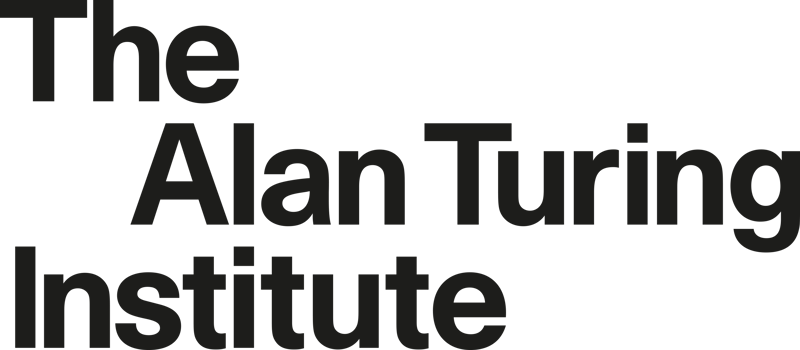}
    };
    }
}

\usepackage{booktabs}
\usepackage{multirow}
\usepackage{xcolor}
\usepackage{colortbl}

\definecolor{lightorange}{HTML}{faa755}
\definecolor{lightblue}{RGB}{220,235,250}

\tcbset{
  takeawaysbox/.style={
    title=Takeaways,
    colback=lightblue!80,
    colframe=black,
    fonttitle=\bfseries\small,
    coltitle=white,
    colbacktitle=black,
    enhanced,
    attach boxed title to top left={xshift=2.5mm,yshift=-2.5mm},
    boxed title style={rounded corners, size=small, colframe=black, colback=black},
    width=\linewidth,
    arc=3.5mm
  }
}

\definecolor{upcolor}{RGB}{56, 142, 60}   
\definecolor{downcolor}{RGB}{211, 47, 47} 
\definecolor{groupgray}{gray}{0.95}       
\definecolor{mybrown}{HTML}{E3D9CC}
\definecolor{mygreen}{HTML}{bcd1ca}
\definecolor{mydarkbrown}{HTML}{d97557}
\definecolor{mydarkgreen}{HTML}{3b8f7a}
\definecolor{myblue}{HTML}{6c9ecc}
\definecolor{mypink}{HTML}{c26584}
\definecolor{coral}{HTML}{EBAF9A}      
\definecolor{peach}{HTML}{F3C9B8}      
\definecolor{mustard}{HTML}{D6B04A}    
\definecolor{marigold}{HTML}{F1B86B}   
\definecolor{slate}{HTML}{92A0A8}      
\definecolor{sky}{HTML}{A7C9E8}        
\definecolor{teal}{HTML}{67B2A8}       
\definecolor{seafoam}{HTML}{B9E3D6}    
\definecolor{olive}{HTML}{A6B182}      
\definecolor{mint}{HTML}{C8EAD7}       
\definecolor{plum}{HTML}{8E5A7A}       
\definecolor{sienna}{HTML}{B87459}     
\definecolor{ochre}{HTML}{C7A35A}      
\definecolor{charcoal}{HTML}{5B6166}   
\definecolor{navy}{HTML}{324A6B}       
\definecolor{lavender}{HTML}{C9B3E0}   

\newcommand{\nUp}[1]{#1\rlap{\ \small$\uparrow$}}     
\newcommand{\nDown}[1]{#1\rlap{\ \small$\downarrow$}} 
\newcommand{\good}[1]{#1\rlap{\ \textcolor{upcolor}{\small\textbf{$\uparrow$}}}}   
\newcommand{\bad}[1]{#1\rlap{\ \textcolor{downcolor}{\small\textbf{$\downarrow$}}}}

\newcommand{\Qwenemoji}{\includegraphics[height=1.1\fontcharht\font`\B]{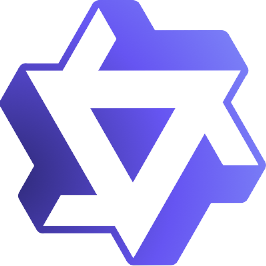}}
\newcommand{\Googleemoji}{\includegraphics[height=1.1\fontcharht\font`\B]{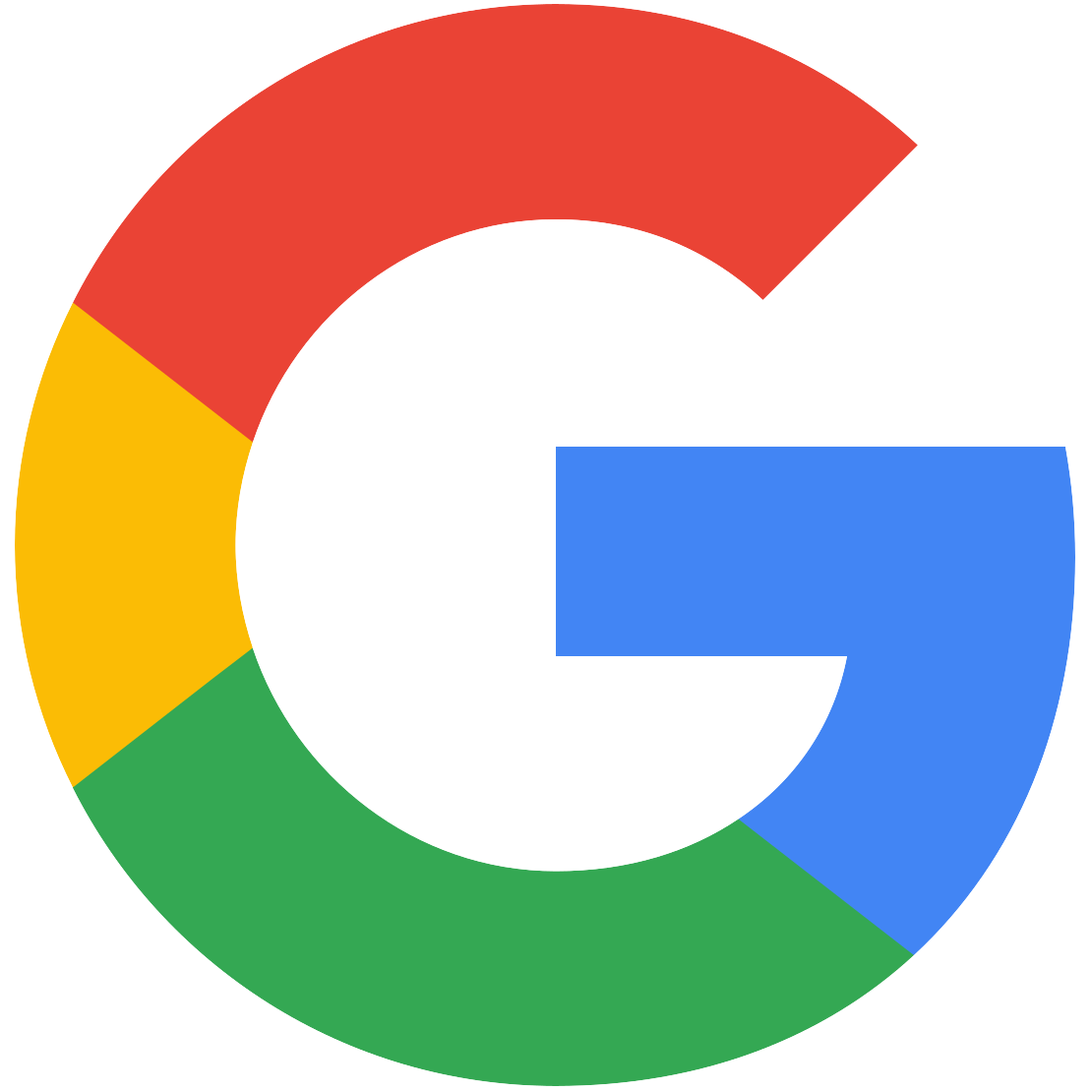}}
\newcommand{\Openaiemoji}{\includegraphics[height=1.1\fontcharht\font`\B]{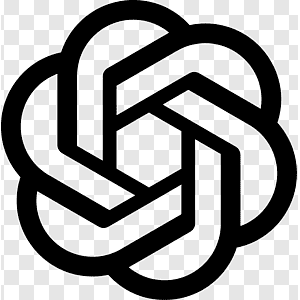}}
\newcommand{\claudemoji}{\includegraphics[height=1.1\fontcharht\font`\B]{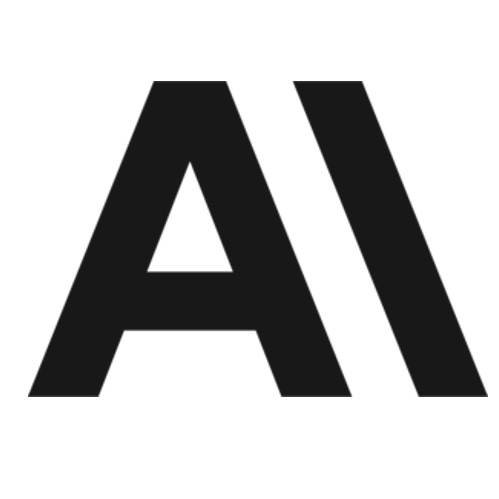}}
\newcommand{\kimimoji}{\includegraphics[height=1.1\fontcharht\font`\B]{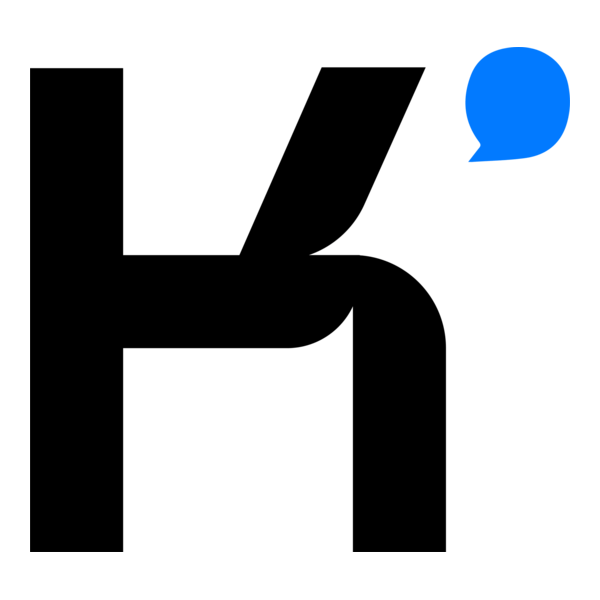}}
\newcommand{\deepseekemoji}{\includegraphics[height=1.1\fontcharht\font`\B]{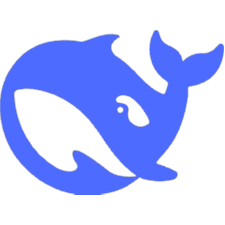}}
\newcommand{\zhipuemoji}{\includegraphics[height=1.1\fontcharht\font`\B]{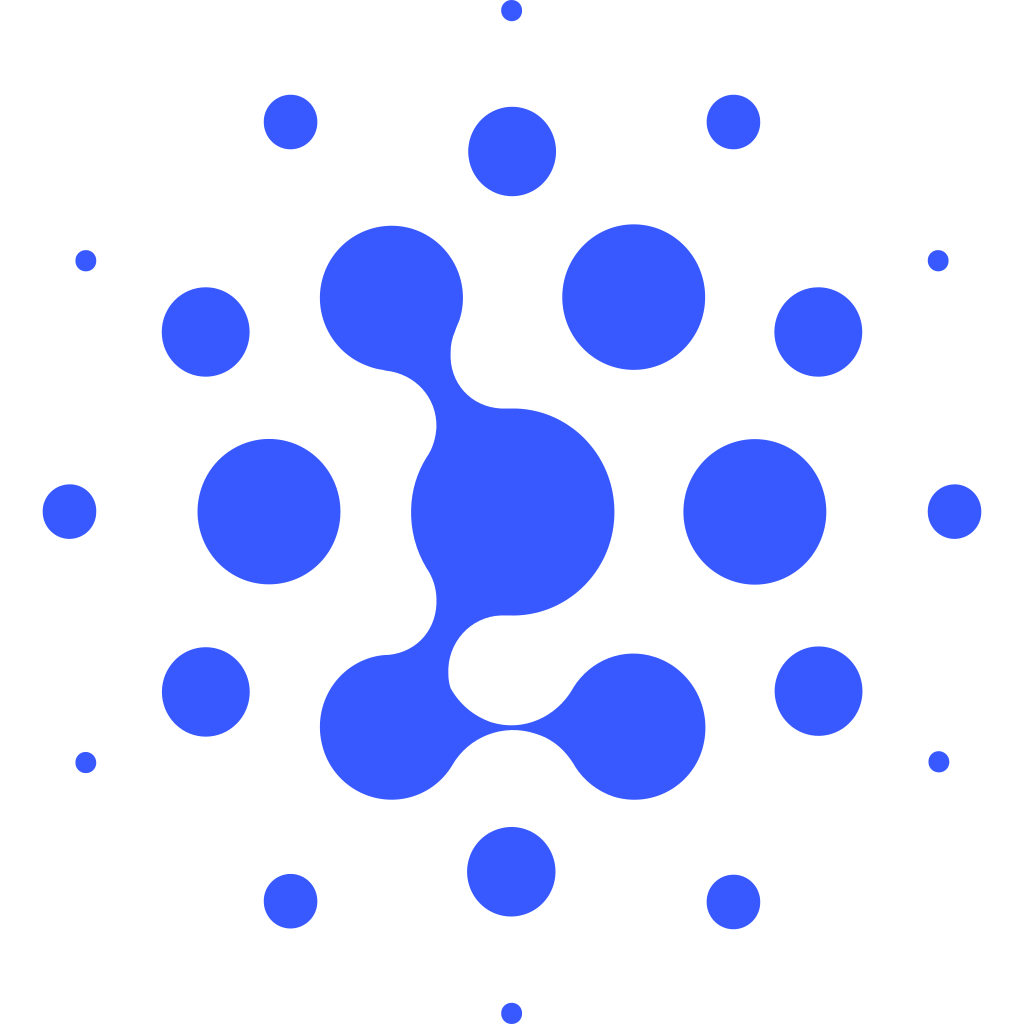}}

\newcommand{\metaemoji}{\includegraphics[height=1.1\fontcharht\font`\B]{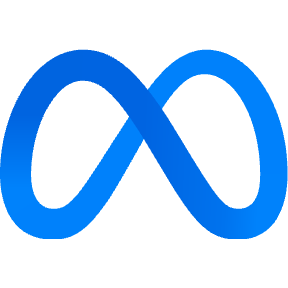}}
\newcommand{\minimaxemoji}{\includegraphics[height=1.1\fontcharht\font`\B]{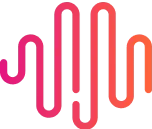}}

\definecolor{myPurple}{RGB}{106,12,173}
\definecolor{myGreen}{RGB}{34, 140, 35}

\usepackage[T1]{fontenc}

\newcommand{\email}[1]{\href{mailto:#1}{\texttt{#1}}}

\usepackage[capitalize,noabbrev]{cleveref}

\theoremstyle{plain}

\theoremstyle{definition}

\theoremstyle{remark}

\usepackage[textsize=tiny]{todonotes}

\icmltitlerunning{Evaluating Deep Data Research on LLMs}

\newsavebox{\teaserbox}
\savebox{\teaserbox}{\includegraphics[width=0.99\textwidth]{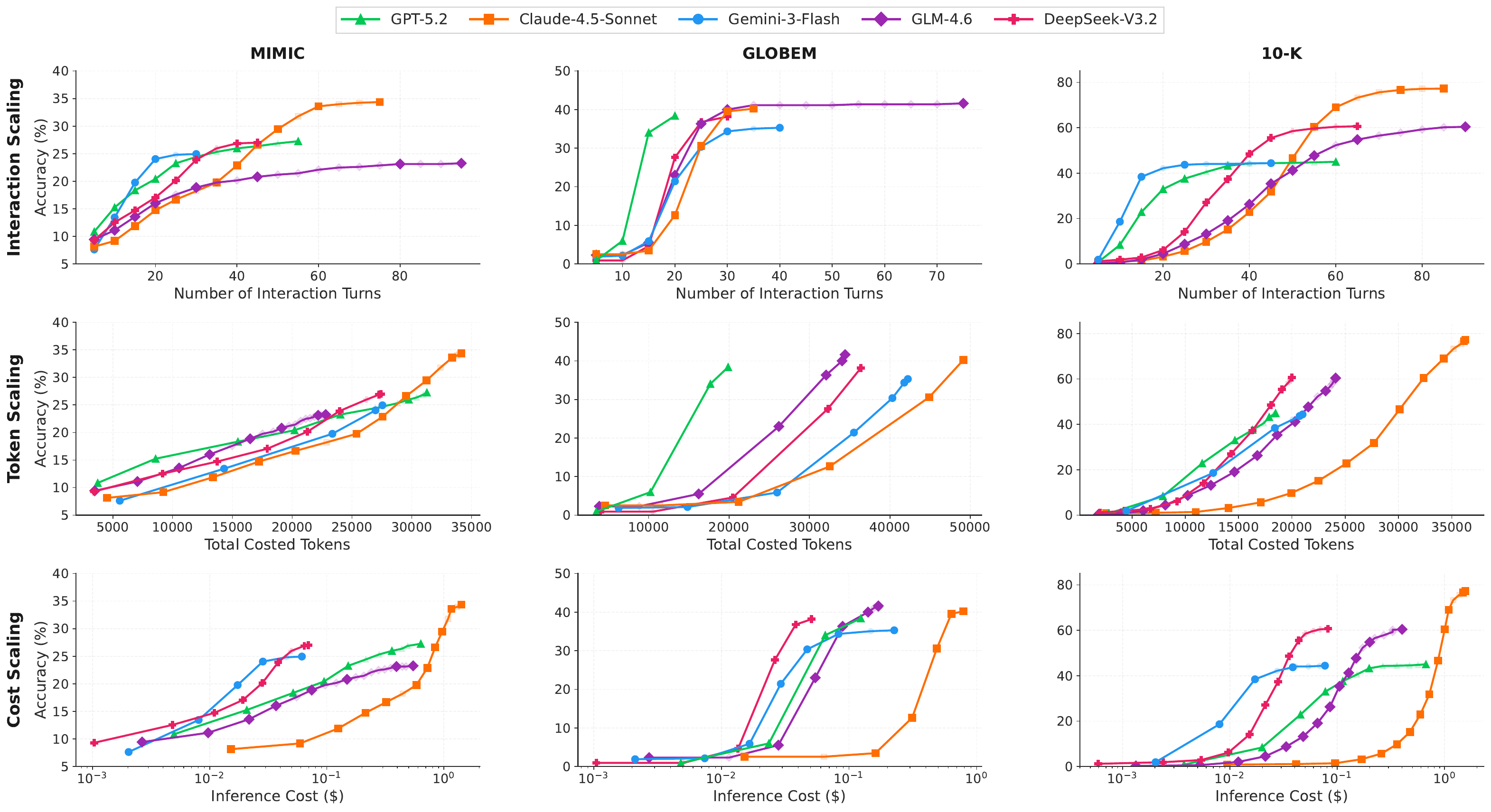}}

\makeatletter
\icml@noticeprintedtrue
\makeatother
\begin{document}

\twocolumn[  
\begin{titlecard}
  {\LARGE\bfseries\sffamily
\raisebox{-0.2\height}{\includegraphics[height=2.0em]{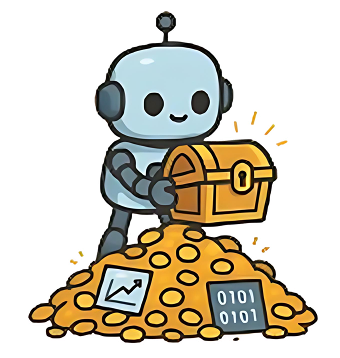}}%
\hspace{0.2em}%
Hunt Instead of Wait: Evaluating Deep Data Research on Large Language Models
}
  
  \vspace{0.3cm}
  
    {\normalsize Wei Liu$^{\spadesuit}$, Peijie Yu$^{\heartsuit}$, Michele Orini$^{\spadesuit}$, Yali Du$^{\spadesuit\clubsuit}$, Yulan He$^{\spadesuit\clubsuit}$}
    
    {\small $^\spadesuit$King's College London, $^\heartsuit$Tencent, $^\clubsuit$The Alan Turing Institute}
  
  \vspace{0.3cm}
  
  \textbf{Abstract:} The agency expected of Agentic Large Language Models goes beyond answering correctly, requiring autonomy to set goals and decide what to explore. We term this \textit{investigatory intelligence}, distinguishing it from \textit{executional intelligence}, which merely completes assigned tasks. Data Science provides a natural testbed, as real-world analysis starts from raw data rather than explicit queries, yet few benchmarks focus on it. To address this, we introduce \textbf{Deep Data Research (DDR)}, an open-ended task where LLMs autonomously extract key insights from databases, and \textbf{DDR-Bench}, a large-scale, checklist-based benchmark that enables verifiable evaluation. Results show that while frontier models display emerging agency, long-horizon exploration remains challenging. Our analysis highlights that effective investigatory intelligence depends not only on agent scaffolding or merely scaling, but also on intrinsic strategies of agentic models.
  
  \vspace{0.3cm}
  
  {\scriptsize \textbf{Project:} \url{https://huggingface.co/spaces/thinkwee/DDR_Bench}}
  
  {\scriptsize \textbf{Correspondence:} 
    \email{wei.4.liu@kcl.ac.uk},
    \email{yulan.he@kcl.ac.uk}
  }
  
  \vspace{0.3cm}

\end{titlecard}

\centering
\usebox{\teaserbox}
\refstepcounter{figure}

\raggedright
\textbf{Figure \thefigure:} Inference-time scaling performance in DDR-Bench across different dimensions. The y-axis reports checklist accuracy. Beyond final accuracy, DDR-Bench provides rich test-time exploration information from different scaling dimensions, enabling detailed analysis of model agency behaviour. See details in~\Sref{sec:invetigatory_dynamics}.
\label{fig:scaling}
\vspace{0.3cm}

]

\section{Introduction}
\begin{figure*}[htbp]
    \centering
    \includegraphics[width=0.99\linewidth]{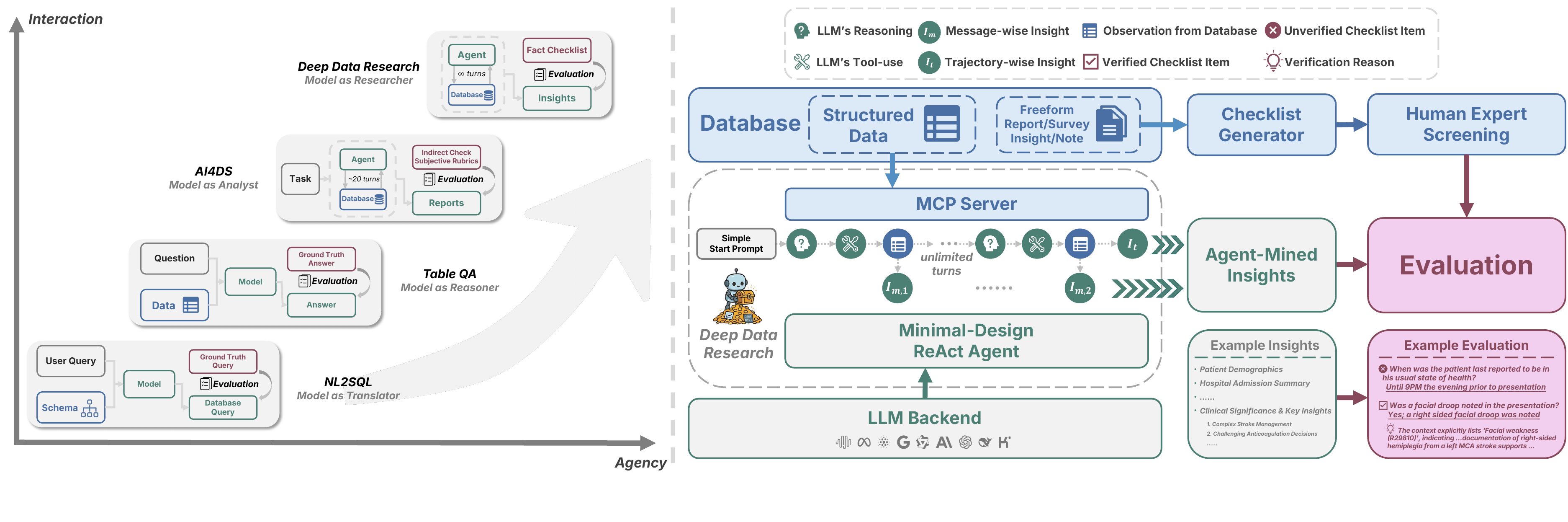}
    \caption{\textbf{Left}: Compared with previous tasks, \textit{DDR} maximises exploration openness and agency, focusing on the direct evaluation of insight quality. \textbf{Right}: Overview of the DDR-Bench. Details of the trajectory samples are shown in Appendix~\ref{appendix:traj_sample}.}
    \label{fig:overall_framework}
\end{figure*}
Agentic large language models (Agentic LLMs)~\cite{zhang2025landscapeagenticreinforcementlearning} extend conventional LLMs~\cite{brown2020language, chatgpt, gpt4, openai2025gpt5} from reactive response to interactive decision making. By integrating models with tools and memory, recent agentic systems are able to complete complex real-world tasks involving long-horizon behaviour.~\cite{plaat2025agenticlargelanguagemodels,wang2025aiagenticprogrammingsurvey,yao2023react,xu2025theagentcompanybenchmarkingllmagents,wang2025odysseybenchevaluatingllmagents,yehudai2025surveyevaluationllmbasedagents}. However, most existing evaluations of agentic LLMs implicitly assume that the task objective or research question is specified in advance. Models are primarily assessed on their ability to execute predefined goals.

This framing conflates two qualitatively different forms of intelligence: \emph{executional intelligence}, which concerns how well an agent carries out a given task, and \emph{investigatory intelligence}, which concerns whether the agent can autonomously decide what is worth investigating. Yet, current evaluation frameworks rarely assess this capability directly.

\textit{AI for Data Science} provides a natural benchmark for this form of intelligence. Human analysts typically begin with structured data, without predefined questions or tasks. They examine the data for irregularities, form tentative hypotheses, and iteratively refine them as new patterns emerge. They will not assume questions or targets before accessing the data. In contrast, much of the existing research on large language models for data science (LLM4DS) remains be evaluating \emph{executional intelligence}, treating a pre-defined, user-posed question as the primary objective for agentic models~\cite{tang2025llmagentasdataanalystsurvey,chen2025largelanguagemodelbaseddata,zhou2025surveyllmtimesdata,sun2025dataagentholisticarchitecture,hong-etal-2025-data}. Recent studies have started to move towards more autonomous data analysis by enabling agentic LLMs to carry out open-ended data explorations and produce high-level insight reports~\cite{zhang2025deepanalyze,sundar2025i2istradainformationinsights}. Evaluation in these settings, however, still largely relies on a combination of low-level correctness checks and subjective assessments of report quality~\cite{nascimento2024llm4dsevaluatinglargelanguage, chen2025scienceagentbenchrigorousassessmentlanguage, zhang2025deepanalyze}, and even without explicit questions, it also includes detailed instructions on what to investigate in the prompt. What's more, such settings are often small in scale, with agents' interactions with databases typically limited to fewer than dozens of steps.

In parallel, deep research benchmarks require agents to perform open-ended web search by coordinating search and browsing tools and generating reports~\cite{wong2025widesearchbenchmarkingagenticbroad,wan2025deepresearcharenaexamllms,zhang2025deepresearchsurveyautonomous}. While these benchmarks also aim to evaluate the agency of agentic LLMs, they operate on unstructured web content and largely limit tool-use to issuing search queries. Moreover, they continue to face significant evaluation challenges, as assessments often depend on subjective rubrics, frequently implemented via LLM-as-a-Judge or on measuring faithfulness to reference websites as a proxy for quality~\cite{du2025deepresearch}.

In this paper, motivated by this gap, we formalise \textbf{\emph{Deep Data Research} (DDR)}, an open-ended setting in which an agent is given only a structured database and a generic toolset, without predefined questions, objectives, or interaction limits. Models are required to autonomously carry out long-horizon tool-use to explore the data, formulate and test hypotheses, decide when to terminate exploration, and ultimately report the insights they uncover. Evaluation is performed based on a checklist derived from the free-form text components of the database, which is used to verify the factual claims in the generated report, supporting an interpretable, objective, and scalable evaluation. Compared with prior work on table question answering~\cite{lu2025large} or report generation in the LLM4DS research~\cite{zhang2025deepanalyze}, DDR enables fully open-ended data exploration while retaining rigorous and verifiable evaluation. To make DDR evaluable at scale, we introduce DDR-Bench, a benchmark that instantiates Deep Data Research over large real-world databases. As shown in Figure~\ref{fig:overall_framework}, DDR-Bench critically challenges models' long-horizon interaction competence and their ability to exhibit sustained, autonomous agency in open-ended investigative settings. Beyond reporting evaluation outcomes, we conduct a systematic analysis of models’ long-horizon interaction patterns, examining from multiple perspectives how agency steers models through the \emph{Deep Data Research} (as shown in Figure~\ref{fig:scaling}). Together with other systematic analyses on model behaviour and module design, we provide a focused empirical examination of current mainstream LLMs, shedding light on the gap between existing agentic AI systems and truly effective investigatory intelligence.
In summary, this paper makes the following contributions:
\begin{enumerate}
    \item We formalise \emph{Deep Data Research} (DDR). This open-ended agentic setting isolates \emph{investigatory intelligence} in structured data environments, requiring models to autonomously explore data, generate and validate hypotheses, and determine when exploration should terminate, without predefined questions or objectives.
    \item We introduce \textbf{DDR-Bench}, the first large-scale benchmark for DDR, which enables fully open-ended and long-horizon data exploration while supporting objective and interpretable evaluation through checklist-based verification of claims grounded in the database.
    \item We systematically evaluate mainstream LLMs on DDR-Bench and reveal persistent limitations in current frontier models. Our findings indicate that progress in agentic LLMs requires not only longer error-free trajectories but also effective use of agency to reliably identify information gaps, adaptively resolve uncertainty during long-horizon exploration, and maintain a globally stable exploration policy.
\end{enumerate}

\section{Method}
\begin{figure*}[htbp]
    \centering
    \includegraphics[width=0.99\linewidth]{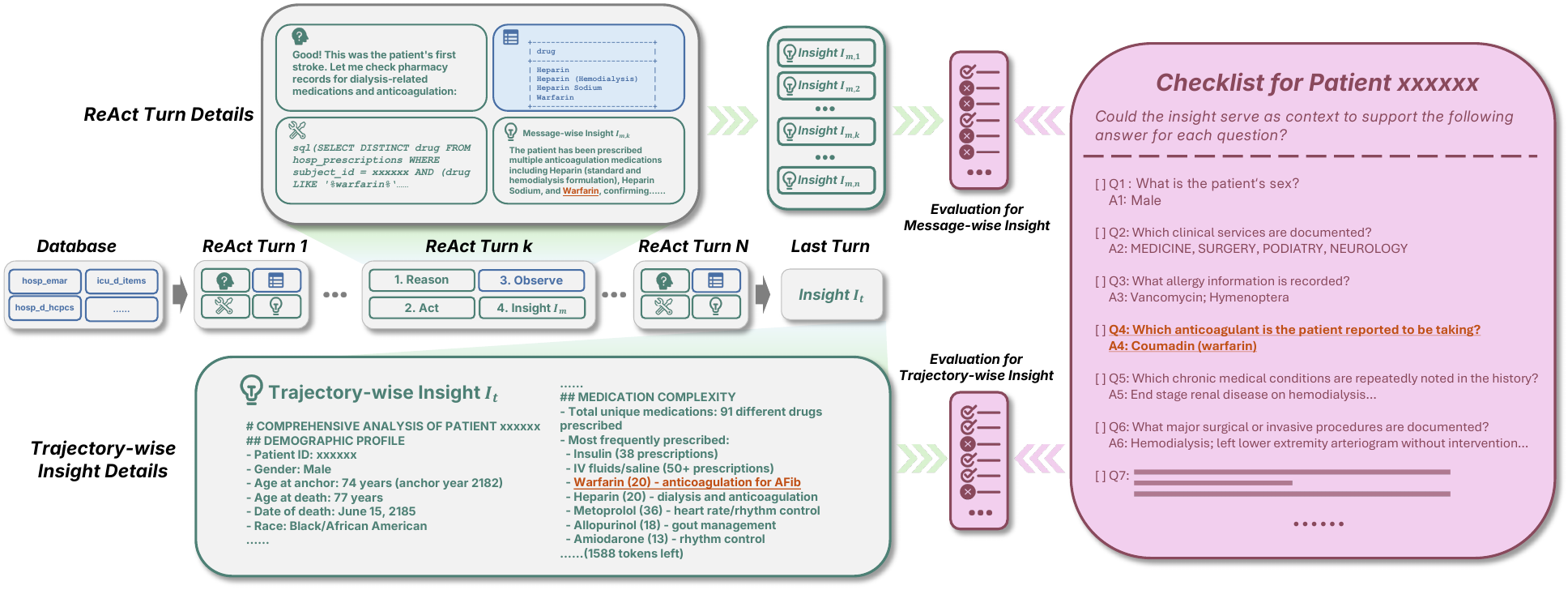}
    \caption{A case of Claude Sonnet 4.5's trajectory and evaluation checklist in the MIMIC scenario of DDR-Bench. Verified fact and supporting insights are \underline{underlined}. See details of this trajectory in Figure~\ref{fig:traj_sample_mimic}. The patient id is anonymised.}
    \label{fig:case}
\end{figure*}
\subsection{Deep Data Research}
As shown in Figure~\ref{fig:overall_framework}, the task of \textit{Deep Data Research} can be formalised as $I = DDR(LLM, D, T)$, where, given a database $D$, an LLM iteratively queries the database using a tool set $T$ (e.g., SQL and Python), with no predefined limit on the number of interaction rounds. The process terminates only when the model judges that sufficient information has been gathered to produce a report comprising multiple insights $I$.
Notably, the LLM is not provided with an explicit question or a predefined objective. Instead, it receives a simple start prompt specifying the \textit{task entity}, which serves as the entry key for database exploration, for example, ``\emph{Start analysing the user with userid=2048}.''
In the first interaction round, the LLM is given basic metadata about the database, including the available tables and a brief description of each. In each subsequent round, the model observes all previous results and then generates reasoning tokens $r$ and tool invocation tokens $t$. Then, these tool calls are executed on the database, and the resulting observations are returned to the model, denoted as $o$.
Through this ReAct-style~\cite{yao2023react} interaction sequence $(r, t, o)$ over multiple rounds, agentic models have access to the full reasoning and action trajectory, with interleaved thinking and execution. 

The model autonomously determines when to stop exploration and produces two types of insights $I = (I_m, I_t)$. The first is \underline{message-wise insight $I_m$}, in which the model is prompted (see Appendix~\ref{appendix:I_m_prompt}) to interpret the $(r, t, o)$ of each ReAct round as an insight paragraph. The second is \underline{trajectory-wise insight $I_t$}, where the model self-terminates and reviews the whole history ${(r_i, t_i, o_i)}_{i=1}^M$ and synthesises it into a single report as the last step of the trajectory, where $M$ is the length of the trajectory.
While the former emphasises process-level insight extraction through incremental interpretations, the latter operates at the outcome level, aggregating global context and applying higher-level reasoning to generate a coherent and comprehensive summary.
This formulation allows the LLM to autonomously set investigative directions and conduct data exploration in a manner analogous to a human data scientist, ultimately generating data insights $I = (I_m, I_t)$. These insights can be viewed as a model-reconstructed context of the task entity within the database. The precise notion of context is database-dependent. For instance, in an electronic health record (EHR) database, the context encompasses the patient's longitudinal healthcare trajectory and health status.

\subsection{Implementation Details of Agent}
We then construct \textsc{DDR-Bench}, a benchmark and evaluation suite for deep data research. It adopts a simple agent scaffold that imposes minimal constraints on model behaviour. Its design is guided by three core principles. First, \textbf{\underline{no query or question is provided}}. The benchmark is not structured as conventional input-output pairs; instead, it consists of a database paired with multiple prompt-checklist instances. Models are given only a minimal start prompt\footnote{A minimal start prompt refers to providing only the task entity, without giving the agent any explicit questions or concrete goals. However, the system prompt still includes basic instructions, such as specifying that the agent should perform data analysis and return results in the ReAct format, as illustrated in~\Sref{appendix:system_prompt}}, such as “\emph{Explore the database and derive insights for \textit{task entity}}”, together with a checklist associated with that entity. Second, \textbf{\underline{the agent framework is deliberately minimal}}. The system prompt follows a lightweight \textit{ReAct} style and excludes explicit workflow, memory, or planning modules. Only two fundamental data analysis tools are exposed through the standard Model Context Protocol~\cite{anthropicMCP2024}, which are SQL and Python. Although complex agent frameworks have demonstrated strong performance on particular tasks, recent advances in agentic LLMs~\cite{zhang2025landscapeagenticreinforcementlearning} suggest that many scaffolded capabilities can be internalised within the models themselves. This motivates the need for benchmarks that provide objective and reliable evaluation signals for intrinsic model capabilities, rather than for auxiliary frameworks. Accordingly, DDR-Bench focuses on assessing model capabilities such as tool-use or long-horizon reasoning, without confounding effects from external scaffolding. Third, \textbf{\underline{exploration is unrestricted}}. No upper bound is imposed on the number of interaction rounds, and termination is determined autonomously by the model.
Together, these design choices maximise model agency and position \textsc{DDR-Bench} as a robust and faithful testbed for evaluating emerging \textit{agentic LLMs}.

\subsection{Data Collection and Construction}
In constructing \textsc{DDR-Bench}, we select three fundamentally distinct scenarios, each built on a large-scale real-world database and characterised by its own unique analytical challenges. All databases in the three scenarios have structured data and unstructured text. This integration enables the derivation of reliable checklists from the unstructured components, which are then used to assess the quality of insights obtained through autonomous exploration of the structured data.To produce reports containing high-quality insights, models need to observe the data, adaptively set investigative goals based on the underlying data patterns, and conduct in-depth analysis and interpretation of the phenomena represented by the data. This process goes well beyond simple data retrieval, aggregation, or question answering. Scenarios in DDR-Bench include:
1) \textbf{\underline{MIMIC-IV}}~\cite{johnson2023mimic}, which is a large-scale deidentified EHR database covering patients admitted to the emergency department or intensive care units at Beth Israel Deaconess Medical Centre in Boston, MA. The \textit{Hosp} and \textit{ICU} modules are retained as structured tables, while the unstructured clinical narratives in the \textit{note} module are used to derive checklists. A total of $100$ patients are uniformly sampled with stratification by note count, ensuring coverage of both simple and highly complex clinical courses. This setting requires LLMs to \textit{reason across multiple tables, perform multi-step analyses to reconstruct complete care trajectories while distilling clinically meaningful insights}.
2) \textbf{\underline{GLOBEM}}~\cite{xu2022globem}, which is a Sport and Exercise Psychology database that combines structured wearable sensor signals with participant surveys on social well-being and mental health. The database contains rich longitudinal records and provides a representative setting for time series analysis. Agents are required to identify temporal patterns across heterogeneous modalities, including Bluetooth, walking activity, Wi-Fi, and sleep signals, and to perform cross-domain reasoning that links behavioural regularities to psychological states. Only daily-level raw data are used, as other wearable data in the database are aggregated or derived from these raw signals. After sampling and filtering, 91 users with non-trivial temporal behaviour patterns are retained as task entities. Although GLOBEM does not include unstructured text, insight-level facts can still be derived from survey responses, such as comparisons of psychological and social well-being before and after the intervention. This transformed data construction challenges whether an LLM \textit{genuinely understands activity data and can reason across domains about participants’ mental states, rather than merely fitting numerical values through simple regression}.
3) \textbf{\underline{10-K}}, which is constructed from annual reports of publicly listed United States companies obtained through the official publicly available SEC API~\footnote{https://www.sec.gov/search-filings/edgar-application-programming-interfaces.}. The structured component comprises XBRL-formatted financial statements, while the unstructured component consists of cleaned textual sections that are closely aligned with the financial data, such as Business Description, Selected Financial Data, and Quantitative and Qualitative Disclosures About Market Risk (Items 1, 6, 7A, 8 and others). A total of $100$ representative companies are selected, each with complex financial statements and more than $5,800$ distinct financial facts. In this domain, LLMs are required to \textit{implicitly build financial models through explorative analysis and synthesise their findings into coherent natural language assessments of firms’ economic conditions}.
Across the three domains, there are 291 task entities in total (shown in Table~\ref{tab:checklist_overview}). \textsc{DDR-Bench} prevents data contamination by separating trajectory generation from evaluation, as no questions are posed to the LLM during data exploration. Data are further anonymised by rewriting column names and metadata without altering their meaning. Empirical analysis shows that hallucinations (see details in Section~\ref{sec:hallucination}), where the models perform poor database interactions yet produce correct insights from memorised knowledge, are extremely rare, and the hallucination ratio and evaluation accuracy are not statistically correlated.

\begin{table}[t]
\centering
\small
\caption{Key statistics of the DDR-Bench. A checklist item denotes a fact that is used to evaluate the quality of model-mined insights.}
\label{tab:checklist_overview}
\resizebox{0.49\textwidth}{!}{
\begin{tabular}{l l c c c c c}
\toprule
\textbf{Database}  & \textbf{Records} & \textbf{Tables} & \textbf{Fields} &  \textbf{Checklist Items} \\
\midrule
$MIMIC$  & 200M+ & 29 & 318 & 774 \\
$GLOBEM$  & 55K+ & 6 & 222 & 435 \\
$10\text{-}K$  & 3M+ & 5 & 5,832 & 849 \\
\midrule
\textbf{Total} & \textbf{203M+} & \textbf{40} & \textbf{6,372} & \textbf{2,058} \\
\bottomrule
\end{tabular}
}
\end{table}

\subsection{Evaluation} 

Evaluating deep data research presents a fundamental challenge because models are expected to generate unconstrained insights through exploratory interaction rather than produce fixed responses to predefined questions. Traditional evaluation paradigms for data exploration and report generation exhibit several shortcomings. First, they rely on constructed questions with corresponding answers and assess correctness, which fails to capture the complex relationships among data, reports, and insights. Second, they embed multiple implicit research objectives within detailed instruction prompts, which undermines the open setting and prevents a proper assessment of model agency in investigative intelligence. Third, they depend on subjective report level scores or indirect factual checks, such as validating code execution results, neither of which directly examines whether the reported insights are faithful to the underlying data.
To overcome these limitations, we adopt a checklist evaluation framework over hybrid structured databases. For each database, verifiable factual statements are extracted from unstructured components using GPT-5 mini~\cite{openai2025gpt5} and organised into a fact checklist. We apply GPT-5 mini and use this checklist to evaluate whether an insight offers sufficient contextual evidence to support each fact (See details in Appendix~\ref{appendix:acc_judge_prompts}). Human expert screening is performed to ensure that the mapping from the data domain to the fact domain is surjective, meaning that every checklist item can be supported by analysing some subset of the database. More than fifty domain experts participate in screening to confirm that each fact is reasonably explorable and inferable from the corresponding data.
Each fact is subsequently further evaluated through both manual verification and LLM-based checking. This process results in three data domains comprising 291 task entities and 2,058 verified checklist items, each linked to one or more queryable database components, ensuring that the checklist is objective, verifiable, and defined at the level of individual samples rather than through global subjective judgements. In evaluation, for open-ended checklists such as MIMIC and 10-K, GPT-5-mini assesses whether collected insights support each fact, and accuracy is computed as the proportion of supported items. For closed-form checklists such as GLOBEM, GPT-5-mini answers checklist questions using the collected insights as context, and the responses are compared against ground truth to compute accuracy.  
Checklist categories and representative examples from each domain are presented in Section~\ref{appendix:checklist_stat} and \ref{appendix:checklist_example}, and an evaluation example is shown in Figure~\ref{fig:case}. By relying on objective checklists rather than subjective rubrics, the evaluation avoids disagreement among human judges and directly targets verifiable data exploration and insight interpretation.

\begin{table*}[ht]
    \centering
    \caption{Benchmarking results. The best results are highlighted in \textbf{bold}. Accuracy is defined as the proportion of checklist items verifiable from the model-mined insights, reported as either sample-averaged (over task entities) or item-averaged (over checklist items).}
    \label{tab:all_databases}
    \newcolumntype{C}{>{\centering\arraybackslash}p{1.1cm}}
    \resizebox{0.99\textwidth}{!}{%
    \begin{tabular}{l|*{6}{C}|*{6}{C}|c}
    \toprule
    & \multicolumn{6}{c|}{\textit{Sample-Averaged Accuracy}} & \multicolumn{6}{c|}{\textit{Item-Averaged Accuracy}} & \\
    \cmidrule(lr){2-7} \cmidrule(lr){8-13}
    \multirow{2}{*}{\textbf{Models}} 
    & \multicolumn{3}{c}{\textit{Message-Wise Insights}} & \multicolumn{3}{c|}{\textit{Trajectory-Wise Insights}}
    & \multicolumn{3}{c}{\textit{Message-Wise Insights}} & \multicolumn{3}{c|}{\textit{Trajectory-Wise Insights}} & \multirow{2}{*}{\textit{Overall Avg.}} \\
    \cmidrule(lr){2-4} \cmidrule(lr){5-7} \cmidrule(lr){8-10} \cmidrule(lr){11-13}
    & MIMIC & GLOBEM & 10-K & MIMIC & GLOBEM & 10-K & MIMIC & GLOBEM & 10-K & MIMIC & GLOBEM & 10-K & \\
    \midrule
    \rowcolor{mydarkbrown!40}
    \multicolumn{14}{c}{\textit{\textbf{Proprietary Models}}} \\
    \midrule
    \claudemoji{}~Claude 4.5 Sonnet & \textbf{36.07} & \textbf{40.13} & \textbf{77.61} & 34.67 & \textbf{38.72} & \textbf{60.58} & \textbf{34.37} & \textbf{40.23} & \textbf{77.27} & 32.95 & \textbf{38.85} & \textbf{61.25} & \textbf{47.73} \\
    \Openaiemoji{}~GPT-5.2          & 28.85 & 38.81 & 44.89 & 32.49 & 38.15 & 41.09 & 27.26 & 38.39 & 44.99 & 30.49 & 38.39 & 41.22 & 37.09 \\
    \Openaiemoji{}~GPT-5.1          & 28.37 & 38.31 & 37.12 & \textbf{35.24} & 35.79 & 44.25 & 26.61 & 37.88 & 37.69 & \textbf{33.59} & 35.63 & 44.76 & 36.27 \\
    \Openaiemoji{}~GPT-5 mini       & 30.02 & 35.86 & 46.82 & 27.86 & 31.54 & 37.12 & 28.81 & 36.09 & 46.35 & 26.36 & 31.72 & 36.77 & 34.61 \\
    \Googleemoji{}~Gemini 3 Flash & 26.58 & 35.60 & 44.82 & 20.78 & 36.74 & 21.24 & 24.94 & 35.29 & 44.41 & 19.51 & 36.78 & 21.08 & 30.65 \\
    \Googleemoji{}~Gemini 2.5 Pro   & 21.51 & 33.77 & 24.48 & 20.00 & 35.62 & 15.57 & 19.51 & 33.79 & 25.68 & 18.48 & 35.40 & 16.14 & 25.00 \\
    \Googleemoji{}~Gemini 2.5 Flash & 16.64 & 29.06 & 8.48 & 23.76 & 28.44 & 16.06 & 14.99 & 28.95 & 8.72 & 22.22 & 28.28 & 16.49 & 20.17 \\
    \Googleemoji{}~Gemini 2.5 Flash-Lite & 17.19 & 26.63 & 19.45 & 17.96 & 24.03 & 9.01 & 16.10 & 26.90 & 19.32 & 17.18 & 24.14 & 9.42 & 18.94 \\
    
    \midrule
    \rowcolor{mydarkgreen!40}
    \multicolumn{14}{c}{\textit{\textbf{Open-Source Models}}} \\
    \midrule
    \deepseekemoji{}~DeepSeek-V3.2  & 28.98 & 38.46 & 60.08 & 30.57 & \textbf{38.46} & \textbf{38.15} & 27.00 & 38.16 & \textbf{60.66} & 28.29 & \textbf{38.62} & \textbf{38.16} & \textbf{38.80} \\
    \zhipuemoji{}~GLM-4.6           & 25.03 & \textbf{41.56} & \textbf{60.31} & 26.15 & 37.60 & 36.02 & 23.26 & \textbf{41.61} & 60.42 & 24.42 & 37.70 & 36.16 & 37.52 \\
    \kimimoji{}~Kimi K2            & \textbf{33.61} & 37.14 & 51.06 & \textbf{30.69} & 37.00 & 30.84 & \textbf{31.65} & 37.01 & 51.24 & \textbf{28.68} & 37.01 & 31.10 & 36.42 \\
    \minimaxemoji{}~MiniMax-M2     & 25.39 & 37.07 & 44.17 & 24.36 & 36.81 & 26.88 & 23.90 & 37.24 & 44.66 & 23.13 & 36.55 & 26.82 & 32.25 \\
    \Qwenemoji{}~Qwen3-Next-80B-A3B    & 18.01 & 35.75 & 44.76 & 21.79 & 33.06 & 30.82 & 16.80 & 35.40 & 45.58 & 20.80 & 32.87 & 31.10 & 30.56 \\
    \Qwenemoji{}~Qwen3-30B-A3B     & 21.67 & 35.73 & 42.44 & 18.75 & 37.25 & 14.38 & 20.03 & 35.63 & 42.33 & 18.22 & 37.01 & 14.13 & 28.13 \\
    \Qwenemoji{}~Qwen3-4B          & 17.97 & 25.99 & 41.13 & 18.68 & 27.55 & 19.76 & 16.67 & 26.21 & 40.94 & 17.18 & 27.59 & 19.91 & 24.97 \\
    \Qwenemoji{}~Qwen2.5-72B       & 15.65 & 28.83 & 27.13 & 16.82 & 25.38 & 13.42 & 14.34 & 28.74 & 27.56 & 15.50 & 25.52 & 14.02 & 21.08 \\
    \Qwenemoji{}~Qwen2.5-14B-1M    & 16.75 & 28.75 & 22.69 & 13.68 & 24.29 & 10.13 & 15.50 & 28.80 & 23.56 & 12.66 & 24.14 & 9.78 & 19.23 \\
    \Qwenemoji{}~Qwen2.5-32B       & 14.12 & 25.67 & 27.07 & 14.40 & 25.88 & 7.90 & 13.05 & 25.82 & 27.53 & 13.18 & 25.98 & 8.12 & 19.06 \\
    \Qwenemoji{}~Qwen2.5-14B       & 15.93 & 25.56 & 18.91 & 14.51 & 26.47 & 10.60 & 14.86 & 25.59 & 19.18 & 13.44 & 26.67 & 10.12 & 18.49 \\
    \Qwenemoji{}~Qwen2.5-7B        & 12.81 & 27.20 & 10.52 & 11.08 & 23.75 & 4.51 & 11.63 & 27.36 & 10.46 & 9.95 & 23.91 & 4.59 & 14.81 \\
    \Qwenemoji{}~Qwen2.5-7B-1M     & 14.61 & 29.34 & 9.42 & 5.95 & 25.68 & 3.91 & 12.85 & 29.00 & 9.71 & 5.30 & 25.29 & 3.65 & 14.56 \\
    \metaemoji{}~Llama3.3-70B      & 10.59 & 23.99 & 9.91 & 5.51 & 21.70 & 2.95 & 9.56 & 23.68 & 9.95 & 5.04 & 21.61 & 3.06 & 12.30 \\
    \bottomrule
    \end{tabular}%
    }
\end{table*}

\section{Benchmark Results}
We benchmark a broad set of proprietary LLMs~\cite{anthropic_claude_4_5_2025, openai2025gpt5, comanici2025gemini25pushingfrontier} and open-source LLMs~\cite{deepseekai2025deepseekv32pushingfrontieropen, 5team2025glm45agenticreasoningcoding, kimiteam2025kimik2openagentic, MiniMaxM2_2025, yang2025qwen3technicalreport, qwen2025qwen25technicalreport, llama3}. We report both accuracy (\Sref{sec:overall}) and novelty (\Sref{sec:novelty}). 

\subsection{Overall Performance}\label{sec:overall}

We report four types of accuracy, combining the average over all task entities (Sample-Averaged) or all checklist fact items (Item-Averaged) with message-wise insights $I_m$ and trajectory-wise insights $I_t$, as shown in Table~\ref{tab:all_databases}. Multiple proprietary and open-source models were evaluated, and to date, only Claude 4.5 Sonnet achieves an average accuracy above 40\%. While leading proprietary LLMs show broadly similar performance across most benchmarks, Claude stands out on DDR-Bench, where its agentic capabilities enable it to surpass models such as GPT and Gemini. Top open-source models, including GLM, Kimi, and Minimax, approach the performance of proprietary LLMs.
These findings indicate that deep data research tasks, which closely resemble real-world exploratory behaviour, remain far from saturated by current models. DDR-Bench reduces test-set contamination, even though many of the underlying databases are likely to contain data previously seen by mainstream LLMs. This is achieved by separating execution from evaluation: during execution, models analyse the data without access to any evaluation questions, while evaluation is performed post hoc using checklists that verify the factual support of the generated insights. As the benchmark contains no question-answer-formatted data, it is inherently resistant to data leakage and training-time overfitting.
Additionally, the results show that model performance does not necessarily align between $I_m$ and $I_t$. While $I_m$ captures the model's immediate analysis after each turn of interaction, $I_t$ reflects its ability to integrate context across the entire trajectory and engage in higher-level global reasoning.

\subsection{Novelty Analysis}\label{sec:novelty}
\begin{figure*}[htbp]
    \centering
    \includegraphics[width=0.99\linewidth]{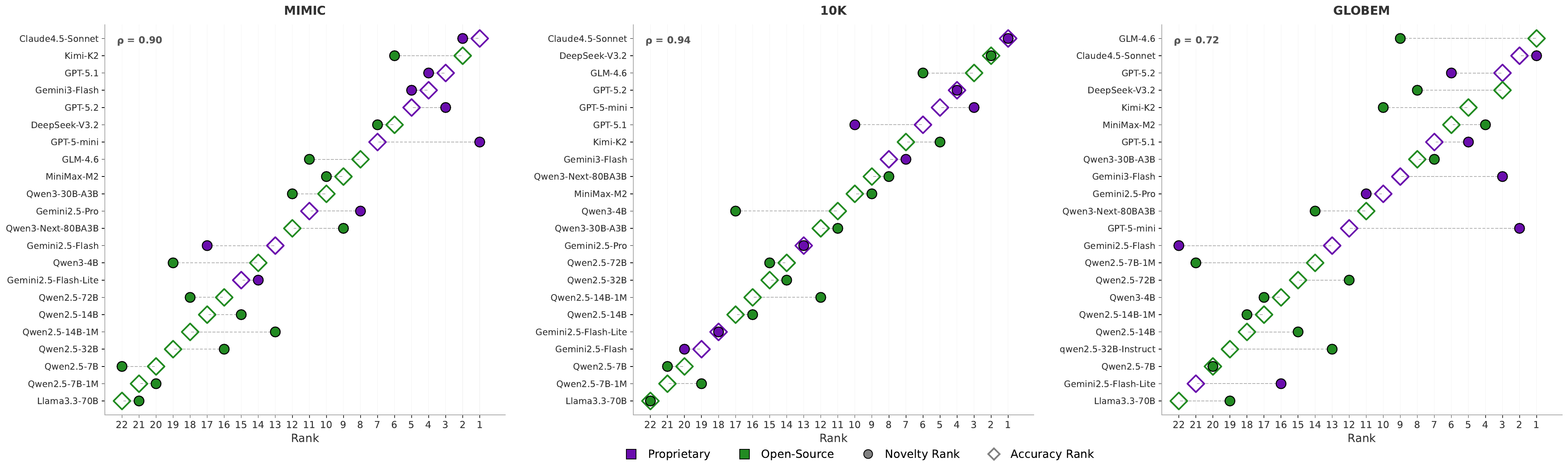}
    \caption{Ranking correlation between novelty and accuracy on {\color{myPurple} Proprietary} and {\color{myGreen} Open-Source} LLMs. Circles denote the novelty rank, and diamonds denote the accuracy rank. Models are ordered by accuracy rank in the figure. All three scenarios present high correlation.}
    \label{fig:novelty}
\end{figure*}

Although DDR-Bench provides a more objective and systematic evaluation than prior open-ended data research benchmarks that rely heavily on subjective LLM-as-a-Judge~\cite{zhang2025deepanalyze, perez2025llm, zhang2025datascibench}, the inherent nature of open-ended tasks inevitably gives rise to false positive cases. A predefined checklist can never exhaustively enumerate all valid or meaningful insights that a model may generate. Manually annotating every output from every evaluation run would be prohibitively expensive and operationally infeasible, and would also introduce substantial subjective bias. Rather than attempting to eliminate such cases, we analyse these false positive novel insights in a pairwise manner.

For each model and each task entity pair, we extract the message-wise insight $I_m$ that are not used in the evaluation of any checklist item and treat them as \emph{novel insights}. Then we employ GPT-5-mini to perform pairwise comparisons between two models on the same task and entity, determining which model produces more useful novel insights (see prompts in~\ref{appendix:novelty_judge_prompts}), with ties allowed. To mitigate position and identity bias, model identities are anonymised, and the order of presentation is randomly swapped. The resulting pairwise comparison outcomes are then aggregated into a global ranking using the Bradley-Terry model~\cite{bradley1952rank}. This novelty-based ranking is subsequently compared with the ranking derived from checklist accuracy. Crucially, we frame novelty evaluation as a pairwise comparison of usefulness rather than a pointwise assessment of novelty. Absolute novelty scores are inherently subjective and difficult to define consistently, whereas relative usefulness judgments are more stable and interpretable. 

As shown in Figure~\ref{fig:novelty}, across all three evaluation scenarios, the ranking induced by novel insight usefulness closely aligns with the ranking based on checklist accuracy. Differences between the two rankings are small, especially among the top-performing models. Further analysis on the number and length of novel insights reveals no systematic relation with these differences. Strong models consistently achieve high checklist accuracy and produce more useful novel insights, regardless of their quantity. This finding highlights a key property of the DDR-Bench evaluation framework. Although checklist-based assessment necessarily covers only a subset of all valid insights, it does not systematically undervalue models that attend to aspects beyond the checklist. Instead, DDR-Bench captures the dominant insight signal. Models with higher checklist scores also tend to generate novel insights that are judged to be more useful.

\section{Investigatory Dynamics}\label{sec:invetigatory_dynamics}
We study \textbf{investigatory dynamics} in this section, which characterise how models behave under agentic test-time scaling in deep data research. This includes how sustained, multi-step interactions influence performance and capability growth (\Sref{sec:scaling}), the structure and evolution of exploration patterns that emerge during extended investigations (\Sref{sec:exploration}), and when models decide to terminate an investigation (\Sref{sec:self_termination}).

\subsection{Scaling Analysis}\label{sec:scaling}

\begin{tcolorbox}[title = \textbf{Takeaway}, colback=takeaway!20, colframe=takeaway!90!Black]

LLMs extract more accurate insights from delaying commitment, and they concentrate reasoning into a small number of highly valuable late-stage interactions. These targeted interactions are built upon longer early exploration.

\end{tcolorbox}

We analysed model performance under test-time scaling using message-wise insights $I_m$. Each interaction round produces some message-wise insights $I_m$, so the quantity of available insights grows and accuracy gradually increases. As no fixed limit is imposed on the number of interaction rounds, leaving the decision of when to stop exploration entirely to the model, the resulting scaling curves also reflect each model’s own assessment of exploration completeness. Some models terminate exploration prematurely despite clear potential for further improvement. The final scaling curves represent the results averaged over all task entities, while the distribution of trajectory lengths (interaction turns) is provided in Appendix~\ref{appendix:turn_distribution}. Test-time scaling is examined from three perspectives: \emph{interaction scaling}, \emph{token scaling}, and \emph{cost scaling}, as shown in Figure~\ref{fig:scaling}.

\paragraph{Interaction Scaling.} Model performance generally follows a sigmoid-shaped trajectory, eventually reaching saturation. While this overall pattern is expected, substantial variation appears in the timing and ceiling of improvement across models. Notably, higher-ceiling models often delay entry into the rapid improvement phase, avoiding early collection of practical information, as seen in Claude, GLM, and DeepSeek. Such patterns resemble a plan-then-act strategy, but examination of trajectories indicates that this behaviour does not arise from explicit planning, since DDR-Bench adopts a minimalism design with no planning prompts; also, the databases are typically too large for comprehensive upfront planning. Instead, it reflects \emph{implicit planning} manifested through interaction dynamics: although no single response explicitly articulates a plan, the sequence of reasoning steps and function calls unfolds as if guided by a predefined plan. Although each action depends on incrementally acquired observations from the database, and is subject to contingencies such as missing values or tool call failures, strong models navigate these uncertainties while preserving the overall exploratory trajectory consistent with an implicit plan.

\paragraph{Token Scaling} Scaling by costed tokens, encompassing all LLM input and output tokens along the exploration trajectory, reveals a different perspective. Token consumption per interaction varies across exploration stages. Under this view, scaling curves shift from a sigmoid shape to one that is initially flat, and then sharply increasing, without clear saturation, highlighting the disproportionately high value of tokens in the final stage. Early broad queries capture readily accessible insights, whereas later-stage gains depend on synthesising accumulated experience and issuing a few highly effective queries. These final tokens signal a shift from breadth-oriented to depth-first exploration. Notably, token consumption is dominated by environmental feedback, specifically, tool execution results from the database. Therefore, the most valuable late-stage tokens are few, indicating that performance gains arise from deep, targeted queries rather than exhaustive search. Analysis of concrete cases shows that top-tier models intensively explore and verify specific hypotheses, issuing complex tool parameters while receiving minimal feedback, often binary responses, since the query or code itself encodes the verification logic. This effect is also indirectly reflected in tool call latency, where the best performing models do not exhibit the longest average tool invocation time, but instead maintain a moderate level (see details in Appendix~\ref{appendix:tool_exec_time}).

\paragraph{Cost Scaling} Model cost is determined by input and output token unit prices, and partly correlated with training and serving expenses. This perspective allows all scaling curves to be fairly compared by accounting for the cost of intelligence. Claude is the most expensive but consistently achieves the highest performance, whereas DeepSeek demonstrates particularly strong cost-effectiveness. A non-uniform horizontal axis illustrates that intelligence gains frequently entail exponential cost increases. Effective model optimisation, through reduced training costs and improved inference-time scaling efficiency, can shift the entire cost scaling curve leftward by multiple orders of magnitude. For example, in the 10-K scenario, both model pairs show similar scaling trends, but at comparable performance levels, Gemini is an order of magnitude cheaper than GPT, and so is DeepSeek compared to GLM. Moreover, efficient models maintain a high rate of effective exploration while producing sufficient exploration steps and insights. This is defined as the valid insight ratio, namely the proportion of interaction rounds that genuinely yield insights among all interactions. See Appendix~\ref{appendix:valid_insight_ratio} for details. 

\subsection{Exploration Patterns}\label{sec:exploration}
\begin{figure*}[t]
    \centering
    \includegraphics[width=0.9\linewidth]{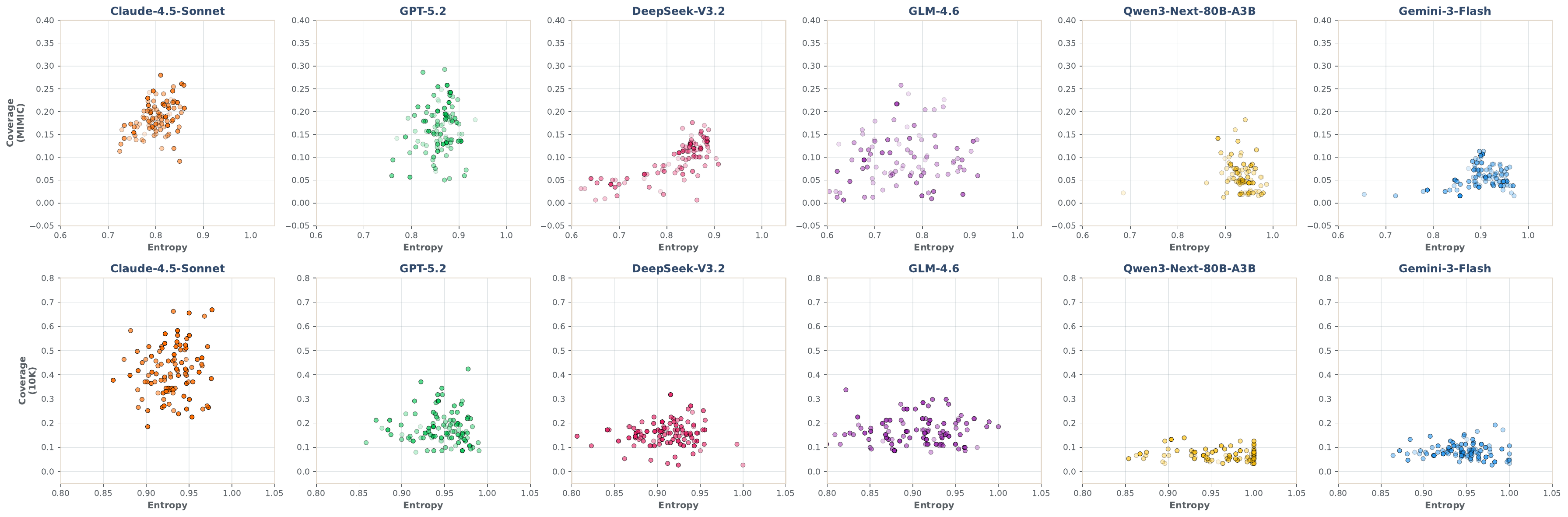}
    \caption{Exploration patterns of different models. The x-axis denotes exploration entropy, reflecting the depth of the model’s search over the database, while the y-axis represents database coverage, indicating the breadth of the search.}
    \label{fig:exploration}
\end{figure*}

\begin{tcolorbox}[title = \textbf{Takeaway}, colback=takeaway!20, colframe=takeaway!90!Black]

Advanced LLMs tend to operate in a balanced exploration regime that combines adequate coverage with focused access. Such a regime is consistently observed across different scenarios.

\end{tcolorbox}
To examine interaction behaviour in greater detail, we visualise model exploration patterns, as shown in Figure~\ref{fig:exploration}. The vertical axis reports \emph{database coverage}, defined as the proportion of distinct fields accessed relative to the total number of available fields. The horizontal axis represents \emph{exploration entropy}, which captures how uniformly a model distributes its access across fields during exploration. To make this quantity comparable across databases of different sizes, we adopt \textit{Normalised Exploration Entropy}. Let $\mathcal{F} = {f_1, f_2, \ldots, f_n}$ denote the set of $n$ distinct fields accessed by a model, and let $c_i$ be the number of times field $f_i$ is accessed. The total number of accesses is $N = \sum_{i=1}^{n} c_i$, yielding an access probability $p_i = c_i / N$ for each field. The entropy of this access distribution is $H = -\sum_{i=1}^{n} p_i \log_2 p_i$ .
We normalise this value by the maximum possible entropy $H_{\max} = \log_2 n$, which is attained under uniform access, obtaining a normalised exploration entropy in the range $(0,1]$:
\begin{equation}
H_{\text{norm}} = \frac{H}{H_{\max}} = \frac{-\sum_{i=1}^{n} p_i \log_2 p_i}{\log_2 n}
\end{equation}
This visualisation jointly characterises exploration breadth and depth. Higher coverage indicates broader exploration across the database, whereas lower entropy reflects more concentrated access patterns, corresponding to deeper, more targeted exploration. Each point represents a single task entity, and point transparency encodes the checklist accuracy achieved for that instance. We omit GLOBEM here since it prioritises algorithm generation, and all models almost always access all fields, which obscures meaningful analysis on exploration patterns from a field-access perspective.

Despite substantial variation in task settings and database scales, most models exhibit remarkably consistent exploration patterns. Claude and GPT, in particular, show balanced behaviour with low variance across instances, suggesting stable internal exploration strategies. In contrast, GLM displays remarkably higher variance, indicating strong heterogeneity across trajectories. For weaker models, the visualisation directly exposes the mechanisms underlying their inferior performance. Qwen and Gemini, for example, tend to access a limited subset of fields, reflecting insufficient exploration breadth, while simultaneously exhibiting high entropy, indicating a lack of focused, selective access even within the fields they explore.
The colour gradient further reveals a clear and robust trend, especially on MIMIC. Instances with balanced exploration patterns that avoid extreme coverage or entropy are consistently associated with higher checklist accuracy across all models. This finding provides empirical support for the implicit planning hypothesis discussed in Section~\Sref{sec:scaling}. Strong LLMs can maintain a coherent and robust internal exploration plan across diverse observations, whereas weaker models fail either by exploring too narrowly or by distributing attention too diffusely.

\subsection{Self-Termination}\label{sec:self_termination}

\begin{figure}[htbp]
    \centering
    \includegraphics[width=0.99\linewidth]{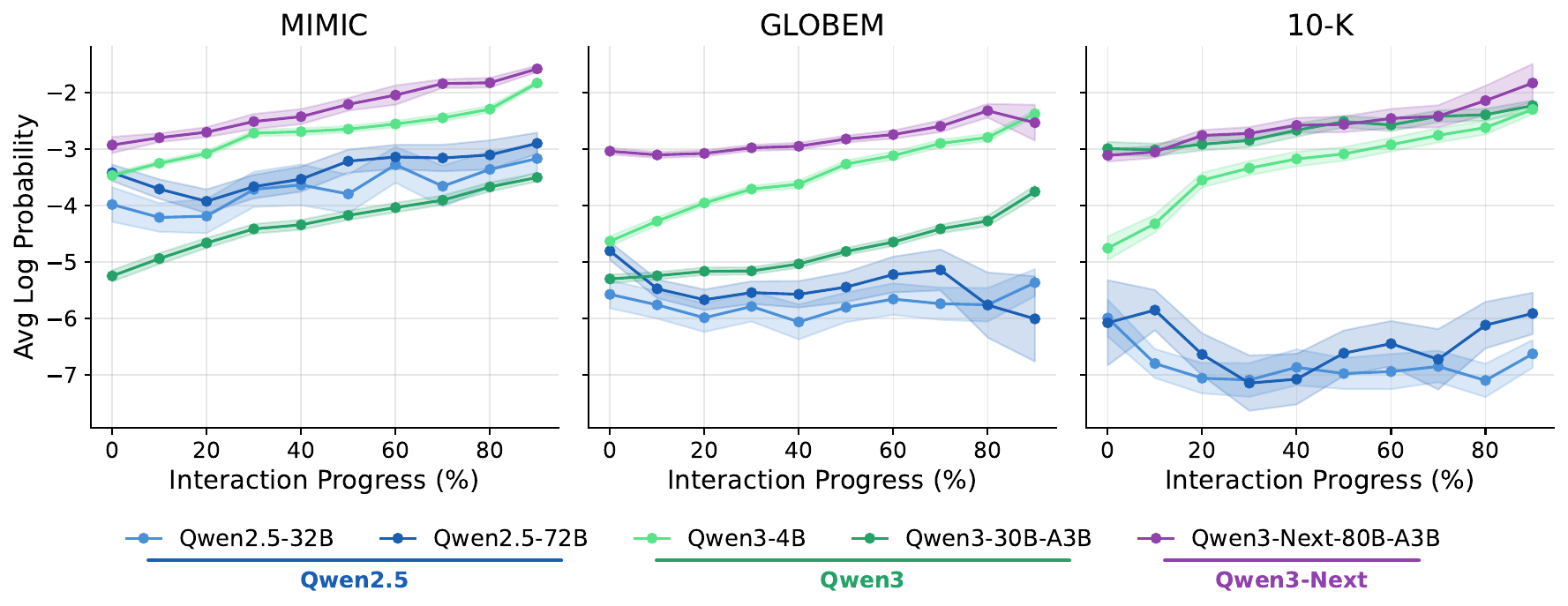}
    \caption{Self-termination visualisation on the Qwen family.}
    \label{fig:self_termination}
\end{figure}

We further examine the model’s self-termination behaviour, namely the point at which an LLM judges that sufficient information has been gathered and exploration should cease. All trajectories generated by the Qwen family are collected, and the probability of directly emitting exploration termination tokens after different numbers of turns is measured as $\frac{1}{N} \sum_{i=1}^{N} \log P(t_i \mid t_1, t_2, \ldots, t_{i-1}, T_{partial})$, where $N$ denotes the length of the finish tokens and $T_{partial}$ is a trajectory prefix containing varying numbers of turns. The results are shown in Figure~\ref{fig:self_termination}. Clear differences emerge across model generations. Qwen3 and Qwen3-Next exhibit a consistently increasing probability, indicating growing confidence that a complete report can be produced as more information is accumulated, whereas the Qwen2.5 series shows pronounced fluctuations and remains uncertain about whether exploration can be terminated at the current step. Moreover, Qwen3-Next maintains higher confidence with lower variance throughout, suggesting that it has more confidence that exploration is progressing towards a more comprehensive and deeper report.

\section{Module Analysis}
Third, we examine how various \textbf{\underline{module}} choices shape behaviour, including training techniques (\Sref{sec:qwen}) and different agent designs (\Sref{sec:agent}).

\subsection{Study on Training Factors}\label{sec:qwen}
\begin{figure*}[htbp]
    \centering
    \includegraphics[width=0.99\linewidth]{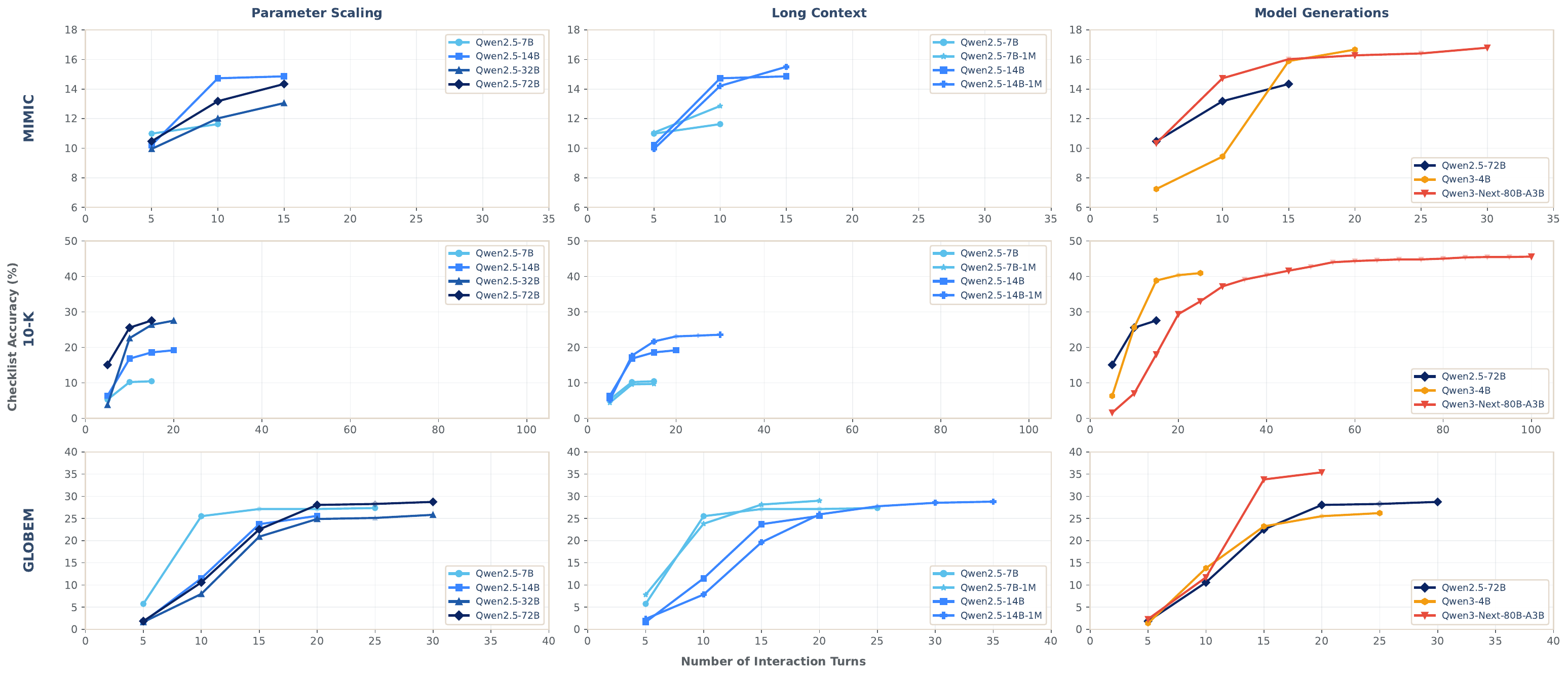}
    \caption{Training-time factors study within the Qwen family. From left to right, the three columns examine inference-time scaling performance across all scenarios for models with different parameter scales, different context optimisation methods, and different model generations with different training strategies.}
    \label{fig:qwen}
\end{figure*}

\begin{tcolorbox}[title = \textbf{Takeaway}, colback=takeaway!20, colframe=takeaway!90!Black]

Scaling is not enough. Meaningful agency require a systematic \emph{agentic-first} training strategy, including targeted pre-training and reinforcement learning.

\end{tcolorbox}

We further analyse how different training-time technical choices would affect the model's investigatory intelligence in the Qwen family~\cite{qwen2025qwen25technicalreport, yang2025qwen3technicalreport}. Qwen offers a wide range of open-source models across multiple versions and scales, which makes it suitable for analysing within a single model family. The results are shown in Figure~\ref{fig:qwen}. 

We first examine the effect of parameter scaling. Increasing model size yields only marginal improvements in accuracy. Even a tenfold increase in the number of parameters results in less than a 3\% gain in final accuracy, and in several settings, smaller models even outperform their larger counterparts. In the case of Qwen2.5 models without explicit agent-oriented training, scaling parameters alone do not enhance proactive exploration and therefore fail to meaningfully strengthen agency.

We then compare the long-context variants of Qwen2.5 at the 7B and 14B scales. Extending the context window does not consistently improve overall performance, nor does it lead to a systematic increase in the number of interaction rounds. Combined with the parameter scaling results above, this suggests that although larger model sizes and longer context lengths are often assumed to facilitate multi-turn agent behaviour, neither factor plays a decisive role in determining agentic capability in deep data research.

Finally, we examine models from different generations, namely Qwen3 and Qwen3-Next, with activated parameter sizes of 4B and 3B, respectively. Despite having fewer activated parameters, both models exhibit a clear increase in the number of exploration rounds on both MIMIC and 10-K. On GLOBEM, the number of exploration rounds remains largely unchanged; however, the models achieve a higher performance ceiling. Taken together, these cross-generation results indicate that improvements in agentic behaviour are primarily driven not by scale, but by a systematic emphasis on reasoning and agentic ability throughout the training pipeline, encompassing both pre-training and post-training stages~\cite{yang2025qwen3technicalreport}.

\subsection{Agent Module Analysis}\label{sec:agent}

\begin{tcolorbox}[title = \textbf{Takeaway}, colback=takeaway!20, colframe=takeaway!90!Black]

Agent modules mainly reshape interaction patterns rather than reliably enhancing insight discovery. Agency in deep data research emerges from stable, implicit coordination between reasoning and open-ended exploration.

\end{tcolorbox}

We conduct an empirical analysis of several agent modules to examine how these capabilities interact with agentic LLMs to maximise effective agency in deep data research.

\paragraph{Reasoning} We increase the reasoning budget of Qwen3-Next-80B-A3B, making it generate more reasoning tokens per interaction round, and evaluate the resulting performance changes. Examination of individual trajectories shows that in many cases the model produces explicit reasoning tokens primarily in the initial rounds, after which it mainly issues tool invocations. Parts of the reasoning are instead implicitly encoded in tool call arguments, such as comments embedded in generated code. This results in an extremely low average number of explicit reasoning tokens, particularly on 10-K and GLOBEM. As reported in Table~\ref{tab:reasoning_comparison}, increasing the reasoning budget substantially raises the average number of reasoning tokens while remarkably reducing the number of interaction rounds. This pattern aligns with expectations, as more detailed reasoning in each round enables more comprehensive data queries, thereby improving efficiency and lowering the total number of interactions. Nonetheless, the final performance metrics show significant fluctuations, indicating a trade-off between reasoning depth and interaction frequency, where neither extreme achieves optimal performance. Information loss can occur when shifting between increased reasoning and increased interaction, suggesting that reasoning and interaction should be treated as dynamically adaptive capabilities of LLMs rather than fixed settings in agent scaffolding.
\begin{table}[h]
\centering
\caption{Comparison of reasoning-token usage, interaction efficiency, and performance under different reasoning budgets for Qwen3-Next-80B-A3B.}
\label{tab:reasoning_comparison}
\scriptsize
\resizebox{0.49\textwidth}{!}{
\begin{tabular}{l|lrrr}
\toprule
\textbf{Stat} & \textbf{Reasoning} & \textbf{10-K} & \textbf{GLOBEM} & \textbf{MIMIC} \\
\midrule
\multirow{2}{*}{Reasoning Tokens Per Turn} 
& Default & 1.20 & 15.26 & 249.75 \\
& Longer & \nUp{357.78} & \nUp{397.11} & \nUp{417.65} \\
\midrule
\multirow{2}{*}{Interactive Turns} 
& Default & 27.93 & 12.93 & 14.56 \\
& Longer & \nDown{11.89} & \nDown{9.41} & \nDown{11.49} \\
\midrule
\multirow{2}{*}{Accuracy - $I_m$ (\%)} 
& Default & 45.58 & 35.40 & 16.80 \\
& Longer & \bad{36.40} & \good{36.78} & \bad{16.67} \\
\midrule
\multirow{2}{*}{Accuracy - $I_t$ (\%)} 
& Default & 31.10 & 32.87 & 20.80 \\
& Longer & \good{37.34} & \good{33.10} & \bad{20.67} \\
\bottomrule
\end{tabular}
}
\end{table}

\paragraph{Memory} DDR-Bench adopts a minimal ReAct Agent design to benchmark a base model’s intrinsic capabilities without interference from external agent frameworks. It operates without any memory mechanism, instead providing the full agent trajectory to the LLM without omission. To examine the effectiveness of memory, comparative experiments on Qwen3-Next-80B-A3B were conducted using a commonly adopted memory mechanism that summarises long trajectories into a local note. This note can be read and updated by the LLM as long-term memory, while retaining only the most recent turns as short-term memory. As shown in Table~\ref{tab:memory}, this long-short-term memory setup makes model behaviour highly unpredictable. Although the note offers a denoised summary and reduces long-context interference, properties that might be expected to encourage more effective exploration, analysis of trajectories reveals that it often induces more aggressive tool-use. In particular, the model tends to read more data per interaction and terminate exploration earlier. This explains several counter-intuitive patterns, such as the reduction in the number of interaction rounds when memory is enabled on 10-K and MIMIC. Moreover, despite access to a summarised history, the agent frequently consumes more tokens overall due to extended exploration within individual rounds and an aggressive data-reading strategy. This accounts for the increased token usage in the 10-K scenario, even when the number of exploration rounds does not increase. Overall, the memory mechanism does not consistently improve final accuracy. While agent frameworks can provide training-free performance gains, their effectiveness is highly sensitive to design choices and typically requires careful, case-specific design.
\begin{table}[h]
\centering
\caption{Effect of a long–short-term memory mechanism on the behaviour and performance for Qwen3-Next-80B-A3B.}
\label{tab:memory}
\small
\renewcommand{\arraystretch}{1.3}
\setlength{\tabcolsep}{10pt}
\resizebox{0.48\textwidth}{!}{
\begin{tabular}{lcccccc}
\toprule
\multirow{2}{*}{\textbf{Scenario}} & \multirow{2}{*}{\textbf{Memory}} & \multicolumn{3}{c}{\textbf{Statistics}} & \multicolumn{2}{c}{\textbf{Accuracy (\%)}} \\
\cmidrule(lr){3-5} \cmidrule(lr){6-7}
& & \multicolumn{1}{c}{Traj Tok.} & \multicolumn{1}{c}{Turns} & \multicolumn{1}{c}{Memory Tok.} & \multicolumn{1}{c}{Traj} & \multicolumn{1}{c}{Msg} \\
\midrule

& -- & 21,256.20 & 27.93 & \multicolumn{1}{c}{---} & 31.10 & 45.58 \\
\multirow{-2}{*}{\textbf{10-K}} & $\checkmark$ & \nUp{24,328.14} & \nDown{18.30} & 2,872.28 & \bad{25.21} & \bad{37.34} \\

\multirow{2}{*}{\textbf{GLOBEM}} & -- & 23,140.60 & 12.93 & \multicolumn{1}{c}{---} & 32.87 & 35.40 \\
& $\checkmark$ & \nUp{26,184.59} & \nUp{20.52} & 2,008.73 & \good{35.86} & \good{35.86} \\

& -- & 19,574.87 & 14.56 & \multicolumn{1}{c}{---} & 20.80 & 16.80 \\
\multirow{-2}{*}{\textbf{MIMIC}} & $\checkmark$ & \nUp{23,650.02} & \nUp{18.28} & 1,670.57 & \bad{14.34} & \bad{15.63} \\
\bottomrule
\end{tabular}
}
\end{table}

\paragraph{Reactive vs. Proactive}

DDR-Bench adopts a query-free, proactive exploration paradigm to assess LLM agency. We additionally implement a conventional reactive variant, in which each checklist item is converted into an explicit user query, resulting in goal-directed and reactive data exploration. Experiments with this variant on Qwen3-Next-80B-A3B, reported in Table~\ref{tab:reactive-comparison}, show a substantial improvement in accuracy in most settings, indicating that the model performs markedly better when provided with clearly specified objectives. This contrast highlights that requiring the model to autonomously identify and prioritise goals poses a significantly more demanding challenge, and thus more directly probes its agentic capabilities. At the same time, the observed gains suggest that, under explicit objectives, the underlying tasks are in principle solvable rather than ill-defined or artificial. Nevertheless, the improvements do not reach full correctness, and performance even degrades in certain cases, reflecting inherent model limitations, as the presence of a clearly defined goal does not guarantee successful task completion.
\begin{table}[h]
\centering
\caption{Comparison of proactive (query-free) and reactive (goal-directed) exploration on Qwen3-Next-80B-A3B.}
\label{tab:reactive-comparison}
\resizebox{0.45\textwidth}{!}{
\begin{tabular}{l|ccc|c}
\toprule
\textbf{Modes} & \textbf{MIMIC} & \textbf{10-K} & \textbf{GLOBEM} & \textbf{Avg} \\
\midrule
\multicolumn{5}{l}{\textit{Proactive(DDR-Bench)}} \\
\quad Message-wise & 16.80 & 45.58 & 35.40 & 32.59 \\
\quad Trajectory-wise    & 20.80 & 31.10 & 32.87 & 28.26 \\
\midrule
\multicolumn{5}{l}{\textit{Reactive}} \\
\quad Overall      & \good{27.13} & \good{70.55} & \bad{31.95} & \good{43.21} \\
\bottomrule
\end{tabular}}
\end{table}

\section{Failure Modes}
In this section, we sampled and manually analysed the failure modes of models. We collect evaluation results from all models across all scenarios, then extract them at the granularity of checklist items. From the items evaluated as incorrectly supported by the model's insight, we randomly sampled instances and linked each item to its corresponding trajectory, $I_t$, and $I_m$. We then manually annotated the reasons why the insights didn't correctly support the fact item. In total, 206 items were annotated. 

We categorise the main errors into eight primary classes, which are discussed below. \textbf{\underline{1) Failure in Exploration}}, such as errors arising from \textit{insufficient breadth} or \textit{insufficient depth} during database exploration. Limited breadth means that the model did not cover the necessary data sources, such as relevant tables or key data fields. It is mainly caused by premature self-termination of the exploration. Limited depth, by contrast, occurs when the model finds relevant data but does not delve deeper and develop more complex queries to extract more complex insights. \textbf{\underline{2) Poor Data-to-Insight Ability}}, which reflects the deficiencies in the pipeline from raw data to statistics and ultimately to insights. The LLMs may stop at descriptive statistics without further interpretation  (\textit{superficial analysis}), draw wrong conclusions from the statistics (\textit{insight misinterpretation}), or, on the contrary, engage in \textit{over-reasoning}, imposing strong assumptions or causal claims that are not supported by the data. \textbf{\underline{3) Lost in the Context}} is particularly prevalent among smaller LLMs, that trajectories are polluted by repetitive or unproductive behaviours, such as repeated \textit{debugging in function calls}, failures in \textit{instruction following}, or making summaries in $I_t$ but missing key points discovered during exploration (\textit{fail in summarisation}).

\begin{figure}[tbp]
    \centering
    \includegraphics[width=0.99\linewidth]{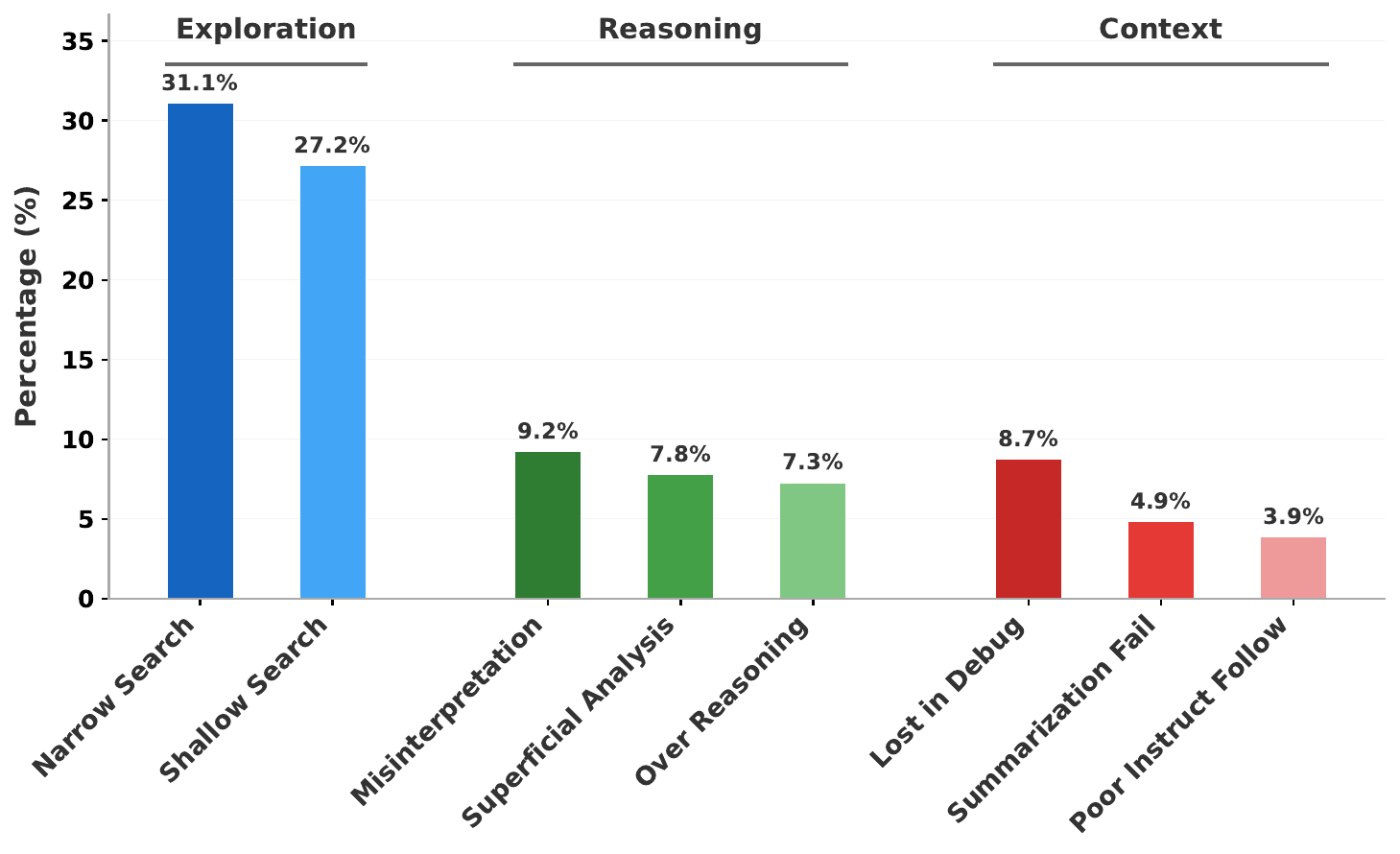}
    \caption{Distribution of manually annotated error types across models and task scenarios.}
    \label{fig:error_dis}
\end{figure}

As shown in Figure~\ref{fig:error_dis}, our findings revealed that 58\% of errors stemmed from insufficient exploration, both in terms of breadth and depth.  This imbalance in exploration often leads to suboptimal results, regardless of the model’s overall capability.
Additionally, around 40\% of the errors were attributed to other factors. For more powerful models, over-reasoning was common, where the model made assumptions not fully supported by the data. In other cases, models misinterpreted the insights, such as mistaking a downward trend for an upward one. Less capable models, on the other hand, tended to make more fundamental errors, such as repeatedly debugging or struggling with missing data, which could disrupt the overall coherence of the analysis.

\section{Hallucination Evaluation}\label{sec:hallucination}
\begin{table}[htbp]
\centering
\small
\caption{Hallucination rates (\%) across models in DDR-Bench, measured as the proportion of insights containing factual but unfaithful information that are not derivable from the provided inputs.}
\resizebox{0.48\textwidth}{!}{
\begin{tabular}{lcccc}
\toprule
\textbf{Model} & \textbf{10-K} & \textbf{GLOBEM} & \textbf{MIMIC} & \textbf{Average} \\
\midrule
Claude 4.5 Sonnet & 4.98 & 1.11 & 6.15 & 4.08 \\
Gemini 2.5 Pro & 2.56 & 0.0 & 0.0 & 0.85 \\
Gemini 3 Flash & 1.82 & 0.0 & 0.0 & 0.61 \\
GLM-4.6 & 3.33 & 1.61 & 0.0 & 1.65 \\
GPT-5.2 & 0.0 & 0.0 & 2.86 & 0.95 \\
GPT-5 Mini & 0.93 & 0.0 & 2.38 & 1.10 \\
Kimi-K2 & 5.56 & 5.71 & 0.0 & 3.76 \\
MiniMax-M2 & 7.69 & 0.0 & 9.09 & 5.59 \\
Qwen3-Next 80A3B & 7.95 & 3.85 & 5.26 & 5.69 \\
DeepSeek V3.2 & 5.30 & 0.0 & 0.0 & 1.77 \\
\bottomrule
\end{tabular}
}
\label{tab:hallucination}
\end{table}

We conducted sampling at the granularity of insights for labelling the possible hallucination, namely, the model generates a factually correct insight with limited or wrong observations from the database. We select from the most recently released models within each model family, as these models are most likely to contain data contamination that could lead to hallucinations. In total, 1,850 insights were randomly sampled for annotation. We pair these insights with all inputs used to produce them, namely the system prompt, model reasoning, function calls, and the database results returned after execution, and checked whether any insight contained facts that were correct but could not be inferred from the available inputs. If a model had memorised relevant information due to data contamination, such cases would manifest as unfaithful but factual hallucinations, thereby distorting benchmarking results. We computed the proportion of insights exhibiting this behaviour, as reported in Table~\ref{tab:hallucination}. 

For most models, the hallucination rate is zero, while for the remaining models it stays below 5\%. Many of these cases involve highly fine-grained entity-level hallucinations that do not affect overall insight judgment or benchmarking outcomes. A common pattern in the 10-K setting is the attribution of financial changes to real-world background events such as acquisitions, even when such information is not supported by the input data. This does not affect evaluation because scoring is based solely on conclusions derivable from the data, and external real-world associations receive no additional credit. Another frequent pattern appears in the MIMIC setting, where medication lists are incomplete due to the use of \verb|LIMIT| in SQL outputs, and models infer additional drugs based on the diagnosed condition and common treatment combinations. While this behaviour likewise does not change evaluation scores, it poses potential risks in healthcare contexts. To mitigate such risks, each insight in DDR-Bench is explicitly linked to its full ReAct trace, including reasoning, tool calls, and execution results. This enables practitioners to verify references for each insight in real-world deployments.

We observe that hallucinations are largely absent in the GLOBEM dataset, likely because its structured and unstructured components originate from distinct domains, namely wearable data and psychological surveys, making cross-domain memorisation less applicable. 
Overall, hallucination rates on DDR-Bench are very low and have minimal impact on benchmarking results. Although larger models may theoretically retain more memorised knowledge, they also exhibit stronger instruction following and adhere more reliably to system prompts that restrict analysis to patterns observable in the provided data.
Moreover, when considered alongside final accuracy, higher hallucination rates do not translate into gains in accuracy. Hallucination rates show extremely weak and non-significant correlations with accuracy (see details in Section~\ref{appendix:hallucination_accuracy_correlation}).

\section{Trustworthiness on the LLM-as-a-Checker}
\begin{table}[htbp]
\centering
\small
\caption{Stability and reliability of LLM-as-a-Checker evaluation in DDR-Bench across repeated runs and scenarios.}
\resizebox{0.48\textwidth}{!}{
\begin{tabular}{lccccccc}
\toprule
\multirow{2}{*}{\textbf{database}} & \multirow{2}{*}{\textbf{Insight}} & \multicolumn{5}{c}{\textbf{Stability}} 
& \multicolumn{1}{c}{\textbf{Reliability}} \\
\cmidrule(lr){3-7} \cmidrule(lr){8-8}
 & & Max & Min & Mean & $\sigma$ & CV & Macro F1 \\
\midrule
\multirow{2}{*}{10-K}
  & $I_m$ & 34.05 & 31.18 & 33.05 & 1.12 & 3.38\% & 92.03\% \\
  & $I_t$    & 15.05 & 13.62 & 14.27 & 0.64 & 4.49\% & 90.05\% \\
\midrule
\multirow{2}{*}{GLOBEM}
  & $I_m$ & 28.79 & 27.27 & 27.88 & 0.83 & 2.98\% & -- \\
  & $I_t$    & 28.03 & 26.52 & 27.42 & 0.63 & 2.31\% & -- \\
\midrule
\multirow{2}{*}{MIMIC}
  & $I_m$ & 17.88 & 16.97 & 17.39 & 0.46 & 2.64\% & 89.88\% \\
  & $I_t$  & 16.47 & 15.00 & 15.47 & 0.57 & 3.71\% & 93.97\% \\
\bottomrule
\end{tabular}}
\label{tab:trustworthiness}
\end{table}
Because DDR-Bench adopts an LLM-as-a-Checker evaluation paradigm, we follow a methodology closely aligned with HealthBench~\cite{arora2025healthbench} and conduct a systematic assessment of its stability and reliability. We emphasise that, in DDR-Bench, the LLM evaluates model-generated insights by matching them against objective ground truth. It does not operate as an LLM-as-a-Judge in settings without ground truth, nor does it rely on subjective, rubric-based criteria. This design choice is intentional and ensures a high degree of objectivity and fairness in the evaluation process.

Specifically, we randomly sample about 10\% of the data across all models and scenarios and repeat the evaluation five times. As reported in Table~\ref{tab:trustworthiness}, the coefficients of variation across all settings are consistently below 5\% for both insight types, demonstrating strong score stability under repeated evaluation. To further assess reliability, we manually annotate sampled trajectories from the MIMIC and 10-K scenarios and compare human annotations with LLM-generated scores. We do not conduct this comparison for GLOBEM, as it involves a closed-form evaluation and does not require an additional reliability check. The resulting macro F1 scores are consistently around 90\%. These results provide strong empirical evidence that the LLM-as-a-Checker employed in DDR-Bench is both stable and reliable.

We further analyse the remaining discrepancies between LLM-as-a-Checker and human judgment. Two primary sources of inconsistency are identified. First, the LLM tends to apply overly strict criteria, requiring insights to explicitly include specific numerical values or statistics. Although these numbers are correctly retrieved by the models, they are often omitted in the final insights because the models are instructed to describe the underlying phenomena rather than enumerate individual figures. Second, in a small number of cases, the LLM produces correct intermediate reasoning but arrives at an incorrect final judgement.

\section{Related Work}
\paragraph{LLM and Agent for Data Intelligence} 
LLMs are increasingly positioned as data analysts and data science agents, reshaping how humans query and manipulate data~\cite{tang2025llmagentasdataanalystsurvey,chen2025largelanguagemodelbaseddata,zhou2025surveyllmtimesdata,sun2025dataagentholisticarchitecture, qiao2025scaling, qiu2026rewarding}. Early work mainly treats data intelligence as answering user specified queries, through table question answering and text to SQL over benchmarks such as WikiTableQuestions, FeTaQA, HybridQA, Spider, and BIRD~\cite{pasupat-liang-2015-compositional,nan2022fetaqa,chen-etal-2020-hybridqa,yu-etal-2018-spider,li2024can}. Subsequent systems move from answers to executable analysis code and visualisations, as in DS 1000 and LLM4DS~\cite{lai2023ds,nascimento2024llm4dsevaluatinglargelanguage}, and introduce agents specialised for plotting and database interaction~\cite{MatPlotAgent2024ACL,xue2024dbgptempoweringdatabaseinteractions,wang2020datashot,pan2025visshepherdconstructingcriticllmbased}. More recent work begins to automate broader data science workflows and proposes agent based benchmarks such as LAMBDA, DataSciBench, and DABStep to evaluate end-to-end behaviour~\cite{li2024autodc,hong-etal-2025-data,zhang2025datascibench,salemi2025llmbasedmultiagentblackboardinformation,egg2025dabstepdataagentbenchmark,dsagent,hollmann2023large,testini2025measuringdatascienceautomation}. Systems like DeepAnalyze and I2I STRADA go further by letting an agentic LLM conduct open-ended workflows over databases and produce narrative reports~\cite{zhang2025deepanalyze,sundar2025i2istradainformationinsights}. Across this line of work, however, either the data problem is still typically posed in advance and evaluation emphasises subjective metrics of report quality, or the report evaluation is transformed into closed-form questions by prompting LLMs. In contrast, DDR-Bench starts from the data alone and asks what an agent can discover, measuring its ability to mine concrete, fact-checked insights rather than simply data statistics.

\paragraph{Proactive Agentic AI} 
Agentic AI studies language model agents that reason, plan, and act through tools and multi-step control, with frameworks such as ReAct and survey work systematising architectures and evaluation~\cite{yao2023react,plaat2025agenticlargelanguagemodels,wang2025aiagenticprogrammingsurvey, Wang_2024,guo2024large,yehudai2025surveyevaluationllmbasedagents}. Within this space, a significant theme is \emph{proactivity}. User-facing agents in visual analytics, mobile interaction, video viewing, and professional workflows monitor activity streams and decide when to intervene to anticipate needs while avoiding unnecessary disruption~\cite{zhao2025proactivevaproactivevisualanalytics,wen2025aiserviceproactiveassistance,yang2025fingertip20kbenchmarkproactive,wang2025proactivevideoqacomprehensivebenchmarkevaluating,lu2024proactive,zhang2024proagentbuildingproactivecooperative}. In these settings, the environment is a user-centred trace. Proactivity is defined as the inference and fulfilment of the user’s immediate goals. From a more abstract perspective, however, a user activity stream is just one kind of data, and “user intent” is just one kind of latent insight that can be inferred from it. A complementary line of work pushes toward open-ended investigation: curiosity inspired and intrinsically motivated methods encourage agents to seek novel states or patterns, and deep research benchmarks ask agents to conduct unconstrained research over the web using search and browsing tools~\cite{dai2025cdecuriositydrivenexplorationefficient,wong2025widesearchbenchmarkingagenticbroad,wan2025deepresearcharenaexamllms,yao2025rigorousbenchmarkmultidimensionalevaluation,zheng2025deepresearcherscalingdeepresearch}. These efforts implicitly adopt a more general notion of proactivity as autonomous insight seeking. Still, their evaluations usually collapse behaviour into global scores for answers or long reports, often using language models as judges. DDR-Bench builds on this generalised view. It treats any data environment, rather than the user alone, as the primary object of investigation, views user intent inference as just one special case of insight discovery, and instantiates this perspective in a setting where agents must decide for themselves what to investigate and are scored by a checklist-based, per-sample fact-checking of the insights they claim to have found.

\paragraph{Open-Ended Data Analysis} Prior work has explored the use of LLMs for open-ended data analysis, yet a principled methodology for benchmark construction remains largely absent. \citet{vykhopen2025beyond} relies on indirect evaluation signals, such as report writing time and qualitative comparisons against a human expert's report, but it does not introduce automated and direct metrics for assessing the quality of extracted insights. \citet{lei2025dacomp} adopts subjective evaluation by LLMs together with multi-level rubrics, and additionally collects baseline reports to compute GSB. However, it still does not directly evaluate insight quality. More importantly, it specifies detailed data exploration objectives for each test case, which means the setting is not truly open-ended data analysis. \citet{sahu2024insightbench} alters the data distribution by manually injecting anomalous patterns as predefined insights to be discovered, resulting in a setup that more closely resembles information retrieval than genuine open-ended analysis. \citet{egg2025dabstepdataagentbenchmark, zhang2025deepanalyze} employ LLM-as-a-Judge and relies on subjective or indirect checklists, such as content relevance, professionalism of organisation and formatting, and the presence of data visualisations. \citet{islam2024datanarrative} similarly uses LLM-as-a-Judge to conduct pairwise comparisons along dimensions such as Informativeness, Clarity and Coherence. \citet{gupta2025bi} automatically generates questions using LLMs, and then applies LLMs as judges for scoring, which again falls short of a fully open-ended setting.
Overall, existing benchmarks exhibit varying degrees of limitation in terms of openness, data scale, and whether insight quality is evaluated directly. Their analyses also concentrate primarily on final accuracy, and therefore cannot provide fine-grained analyses of exploration trajectories in the way that DDR-Bench does.

\section{Conclusion}
DDR-Bench establishes a rigorous framework for evaluating investigatory intelligence, revealing that the transition from reactive execution to proactive discovery relies less on external scaffolding or simple parameter scaling, and more on intrinsic exploration strategies that balance breadth with targeted reasoning. Our findings highlight a critical distinction between executional and investigatory capabilities, observing that effective agency emerges from implicit planning and the autonomous determination of termination criteria rather than rigid workflows. As the field advances, this underscores the necessity of moving beyond query-response alignment toward training paradigms that cultivate end-to-end autonomy, enabling agentic LLMs to not only answer but proactively define goals and pursue them to derive useful insights from large-scale data.

\section{Impact Statement}

This paper introduces DDR-Bench to advance the field of autonomous data science and investigatory intelligence.

\paragraph{Data Compliance and Privacy.}
Our benchmark integrates diverse real-world datasets. We strictly adhere to all relevant data use agreements and ethical regulations, particularly for the medical subset, which utilises fully de-identified data accessed under credentialed requirements. No personally identifiable information is processed, ensuring compliance with standard privacy protocols.
To ensure openness and transparency, we provide the complete code for running the agents and performing evaluation in the supplementary materials, and we will release this code publicly in the future. Sensitive datasets such as MIMIC and GLOBEM will be made available securely via their hosting platform, PhysioNet, in the form of derived datasets, with access granted only to researchers who have completed the required security certification.

\paragraph{Reliability and Traceability.}
We conducted a comprehensive analysis of model hallucinations, finding them to be low in our experimental settings (see \Sref{sec:hallucination}). To further mitigate risks associated with autonomous analysis, DDR-Bench is designed with \textbf{full traceability} as a core principle. Every generated insight is explicitly linked to a specific analysis turn, mapping the claim directly to the executable query (SQL/Python) and the raw environment observation. This structure ensures that all agent-derived conclusions are verifiable and grounded in factual evidence.

\paragraph{Societal Implications.}
While this work aims to democratize data science, we acknowledge the risk of automation bias. These agents are designed to assist human experts by handling laborious exploration, not to replace human judgment in high-stakes decision-making.

\section*{Acknowledgments}
\label{sec:acknowledgments}

This work was supported in part by the UK Engineering and Physical Sciences Research Council through a Turing AI Fellowship (grant no. EP/V020579/1, EP/V020579/2) and the Prosperity Partnership scheme (grant no. UKRI566). Wei is supported by a PhD studentship provided by King's College London (KCL). The authors acknowledge the use of Computational Research, Engineering and Technology Environment (CREATE) at KCL, and Inkfish through the EMBRACE research programme.

\clearpage

\bibliography{example_paper}
\bibliographystyle{icml2026}

\newpage
\appendix

\setcounter{figure}{0}
\renewcommand{\thefigure}{A\arabic{figure}}
\setcounter{table}{0}
\renewcommand{\thetable}{A\arabic{table}}

\onecolumn

\section{Checklist Statistics}\label{appendix:checklist_stat}
\begin{figure*}[htbp]
    \centering
    \includegraphics[width=0.99\linewidth]{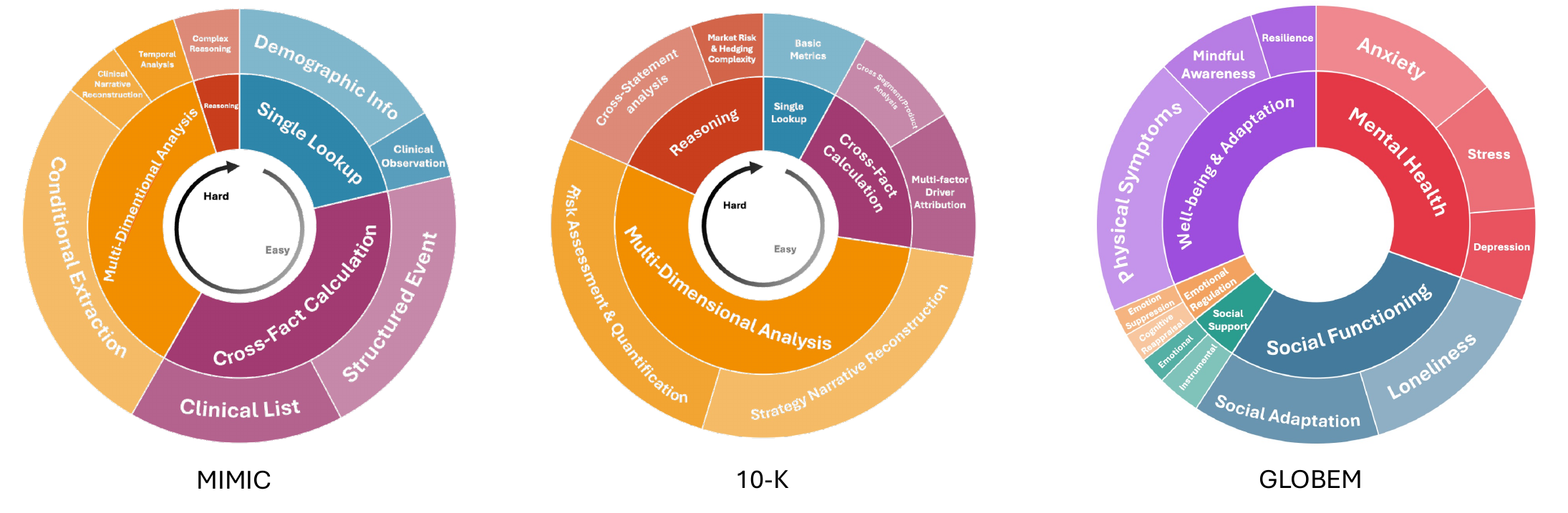}
    \caption{Distribution of checklist items.}
    \label{fig:checklist_distribution}
\end{figure*}
Figure~\ref{fig:checklist_distribution} shows the distribution of checklist items across the three DDR-Bench scenarios. For MIMIC and 10-K, the checklists are free-form. The outer ring groups items by question type, while the inner ring categorises the data analysis capabilities required of LLMs. Simple items, such as demographic information extraction in MIMIC or basic metric extraction in 10-K, require only a single lookup that can be completed with a straightforward SQL SELECT, without complex instructions or computation. In contrast, the most demanding items, such as analysing a patient’s pathology by integrating multiple surgeries, diagnoses, and medications across departments, or assessing market risk by combining ten year financial trends with recent capital flows, require multi-table reasoning at the highest level of complexity. For GLOBEM, the checklist items are closed-formed. Numerical responses are extracted from each user’s surveys, and questions are constructed by comparing changes in the same items before and after the wearable experiment, for example, whether the user’s depression status improved after the experiment. The answers are closed-formed, with three options indicating improvement, no improvement, or no substantial change. When sufficient insights are derived from the wearable data, the LLM checker can answer these trend questions correctly based on the inferred evidence. This temporal comparison avoids pointwise prediction of absolute survey scores for social well-being or psychological status, which are highly subjective and exhibit large variance, and also avoids framing the task as numerical regression, which is not the aim of insight report generation. For GLOBEM, the outer and inner rings correspond to major and minor categories of survey questions. Required data modelling capabilities are not shown in the inner ring, as there is no ground truth modelling approach, and LLMs may flexibly analyse wearable data using Python code and any suitable algorithms.

\section{Agent Framework Ablations}

We intentionally simplify the agent to evaluate model capabilities in a clean testbed; optimising agent scaffolding for high scores would confound assessment. DDR-Bench supports flexible tool extensions via the Model Context Protocol (MCP) for future agent benchmarking.

We presents agent ablations, where memory modules often destabilise performance. Table~\ref{tab:agent_ablations} provides additional experiments on Qwen3-30B-A3B, Qwen3-4B, and GPT-5-mini across three framework configurations: Planning (Plan-and-Execute~\cite{wang2023plan}), Memory (CoALA~\cite{sumers2023cognitive}), and Multi-agent (AutoGen~\cite{wu2024autogen}). Results reveal that complex agents mostly degrade performance relative to the ReAct baseline, except for minor planning benefits in certain settings. Analysis suggests that complex frameworks affect model confidence, leading to premature self-termination or over-/under-thinking.

\begin{table}[htbp]
\centering
\caption{Performance comparison of agent framework configurations across datasets and models. Scores are reported for ReAct (baseline), Plan-and-Execute (+Plan~\protect\cite{wang2023plan}), CoALA (+Memory~\protect\cite{sumers2023cognitive}), and AutoGen (+Multi-Agent~\protect\cite{wu2024autogen}). Higher is better.}
\label{tab:agent_ablations}
\begin{tabular}{llrrrr}
\toprule
\textbf{Dataset} & \textbf{Model} & \textbf{ReAct} & \textbf{+Plan} & \textbf{+Memory} & \textbf{+Multi-Agent} \\
\midrule
\multirow{3}{*}{MIMIC}
  & Qwen3-4B        & 16.67 & 8.14  & 11.46 & 4.44  \\
  & Qwen3-30B-A3B   & 20.03 & 12.27 & 13.57 & 9.04  \\
  & GPT-5-mini      & 28.81 & 23.67 & 22.22 & 12.66 \\
\midrule
\multirow{3}{*}{10-K}
  & Qwen3-4B        & 40.94 & 14.25 & 17.43 & 26.50 \\
  & Qwen3-30B-A3B   & 42.33 & 47.59 & 37.10 & 31.80 \\
  & GPT-5-mini      & 46.35 & 49.82 & 45.35 & 30.04 \\
\midrule
\multirow{3}{*}{GLOBEM}
  & Qwen3-4B        & 26.21 & 22.76 & 23.45 & 22.30 \\
  & Qwen3-30B-A3B   & 35.63 & 25.75 & 22.30 & 23.91 \\
  & GPT-5-mini      & 36.09 & 28.05 & 28.72 & 25.64 \\
\bottomrule
\end{tabular}
\end{table}

\section{Checklist Example}\label{appendix:checklist_example}
Figure~\ref{fig:checklist_example_mimic}, \ref{fig:checklist_example_10k} and \ref{fig:checklist_example_globem} each present a checklist sample from one of the three scenarios in DDR-Bench, namely MIMIC, 10-K, and GLOBEM. MIMIC and 10-K involve open-ended question answering, whereas GLOBEM adopts a multiple-choice format with three fixed options: better, worse, and remained the same. Notably, for open-ended questions, we do not ask a judge LLM to first answer the question based on the agent's collected insights and then compare the answer with the ground truth. Instead, as illustrated in Figure~\ref{fig:case}, we provide the checker LLM with the insights, the question, and the ground truth, and ask it to assess whether the insights can support the given ground truth. This design serves two purposes. First, it avoids the difficulty of normalising open-ended answers for direct comparison. Second, it shifts the focus to whether the collected insights substantively support the underlying facts, rather than relying on exact matching of surface descriptions.
\begin{figure*}[htbp]
\begin{tcolorbox}[title={Checklist Example in the MIMIC Scenario}]
Q1: What is the patient's sex?

A1: Female

\

Q2: Which clinical services are documented?

A2: OBSTETRICS/GYNECOLOGY and MEDICINE

\

Q3: What allergies are listed?

A3: Shrimp; Ondansetron

\

Q4: Was a laparoscopic left salpingo-oophorectomy performed?

A4: Yes. A laparoscopic left salpingo-oophorectomy (LSO) is documented.

\

Q5: Was a single-balloon enteroscopy performed or planned?

A5: Yes. Single-balloon push enteroscopy is documented as a major invasive procedure.

\

Q6: Was pelvic washings performed during surgery?

A6: Yes. Pelvic washings are documented (conversion to ex-lap with LSO and pelvic washings).

\

Q7: Did the patient undergo an ovarian detorsion?

A7: Yes. Laparoscopic ovarian detorsion (left detorsion) is documented.

\

Q8: What did the final pathology of the ovarian tissue show?

A8: Adenocarcinoma involving ovarian tissue.

\

Q9: What diagnosis is listed as the primary oncologic diagnosis?

A9: Metastatic gastric adenocarcinoma

\

Q10: What histologic subtype of gastric cancer is mentioned?

A10: Signet ring gastric adenocarcinoma

\end{tcolorbox}
\caption{Checklist Example in the MIMIC Scenario.}
\label{fig:checklist_example_mimic}
\end{figure*}

\begin{figure*}[htbp]
\begin{tcolorbox}[title={Checklist Example in the 10-K Scenario}]
\small
Q1: How has regulatory risk affected AEP’s ability to recover its costs through rates?

A1: Regulatory risk has remained elevated recently — if regulators do not authorize timely rate adjustments, AEP could face reduced cost recovery and materially lower earnings (significant downside risk to results).

\

Q2: How does regulation affect AEP’s rates and recovery of costs?

A2: Regulation remains central: rates are largely cost-of-service or formula-based and include mechanisms (fuel/purchased power recovery, trackers, formula rates) that enable more timely cost recovery; both state commissions and the FERC play active oversight roles, and recent management focus has been on alternatives to reduce regulatory lag.

\

Q3: What were the main drivers of the company's improved earnings in 2024?

A3: Earnings increased materially in 2024 driven mainly by favorable IRS PLRs that changed retail NOLC ratemaking, supportive rate proceedings, higher transmission-related revenues and increased sales volumes; these gains were partly offset by higher operating costs and certain unfavorable prior regulatory rulings. The improvement was concentrated in the 2024 year.

\

Q4: How did AEP's profitability change in 2024 and what were the main drivers?

A4: Profitability improved notably in 2024 versus 2023, driven mainly by favorable IRS PLRs treating NOLCs, positive rate case outcomes and higher transmission and sales volumes; these gains were partly offset by CCR-related costs, severance and a Texas refund provision. The change was material for the year 2024 compared to 2023.

\

Q5: What trend is described regarding regulatory scrutiny and proceedings of AEP’s utility operations?

A5: Regulatory scrutiny and formal proceedings have increased or remained active in recent years, with regulators probing prudence of costs, past capital expenditures and rate filings, raising the risk of disallowances or adjustments (moderate to significant regulatory pressure).

\

Q6: How is AEP’s transmission business trending and what major initiatives are underway?

A6: AEP’s transmission business is growing materially, driven by substantial multibillion-dollar investments through State Transcos and transmission joint ventures; management has been accelerating construction and filing forward-looking formula rates, with significant projects planned through the next several years.

\

Q7: How did profitability margins trend during the recent period?

A7: Profitability margins improved in 2024 (notably higher net income attributable to common shareholders year-over-year) due to one-time tax/regulatory benefits and operational revenue gains, although margins faced pressure from increased operating costs like CCR compliance and severance charges.

\

Q8: How have regulatory and tax accounting developments affected AEP's balance sheet in the recent period?

A8: Regulatory and tax changes materially reshaped the balance sheet in 2024 — AEP recognized regulatory assets for NOLC revenue requirements, remeasured Excess ADIT (reducing regulatory liabilities), and recorded related deferred tax impacts, producing noticeable shifts in deferred tax and regulatory account balances. These were significant in 2024 as the PLRs and rate filings were implemented.

\

Q9: How does regulatory lag in rate cases affect AEP’s earnings trend?

A9: Regulatory lag from lengthy base rate proceedings tends to depress or delay recovery of costs, causing earnings to be lower or more volatile in the near term (noticeable impact over the most recent rate cycles).

\

Q10: What is the role and recent trend of the Generation \& Marketing segment?

A10: The Generation \& Marketing segment operates competitive retail supply and wholesale trading businesses that face market-driven volatility; it remains an important but more market-exposed part of AEP’s portfolio, and the company has recently narrowed the segment through asset sales while maintaining a sizable retail customer base.

\end{tcolorbox}
\caption{Checklist Example in the 10-K Scenario.}
\label{fig:checklist_example_10k}
\end{figure*}

\begin{figure*}[htbp]
\begin{tcolorbox}[title={Checklist Example in the GLOBEM Scenario}]
Q1: How did the user's emotional support given change during the experiment?

A1: It improved.

\

Q2: How did the user's emotional support received change during the experiment?

A2: It improved.

\

Q3: How did the user's depressive symptoms change during the experiment?

A3: It worsened.

\

Q4: How did the user's anxiety levels change during the experiment?

A4: It worsened.

\

Q5: How did the user's social adaptation change during the experiment?

A5: It remained almost the same.

\end{tcolorbox}
\caption{Checklist Example in the GLOBEM Scenario.}
\label{fig:checklist_example_globem}
\end{figure*}

\section{Interaction Turn Distribution}\label{appendix:turn_distribution}
\begin{figure*}[htbp]
    \centering
    \includegraphics[width=0.99\linewidth]{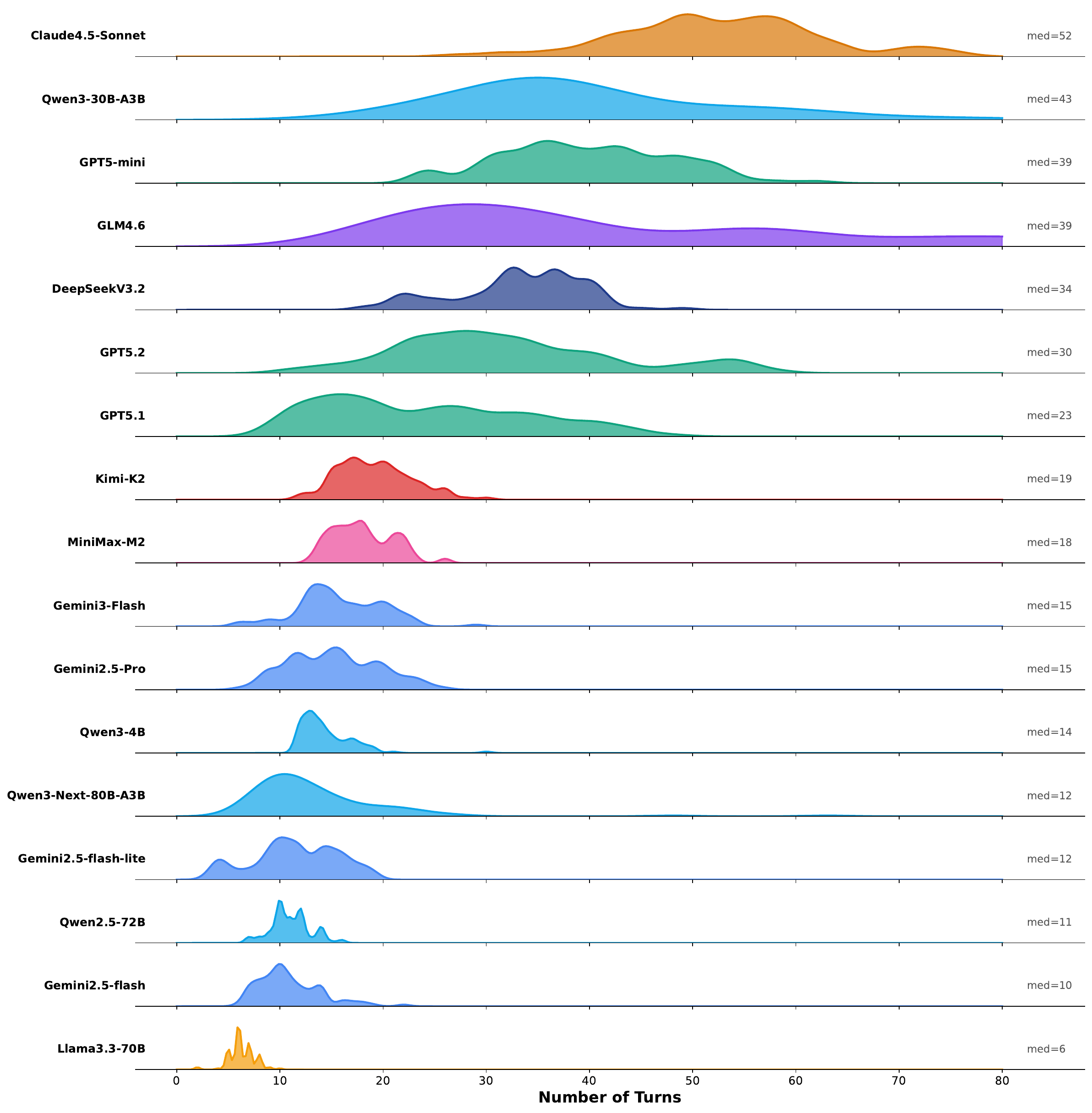}
    \caption{Interaction turn distribution on MIMIC scenario.}
    \label{fig:turn_distribution_mimic}
\end{figure*}

\begin{figure*}[htbp]
    \centering
    \includegraphics[width=0.99\linewidth]{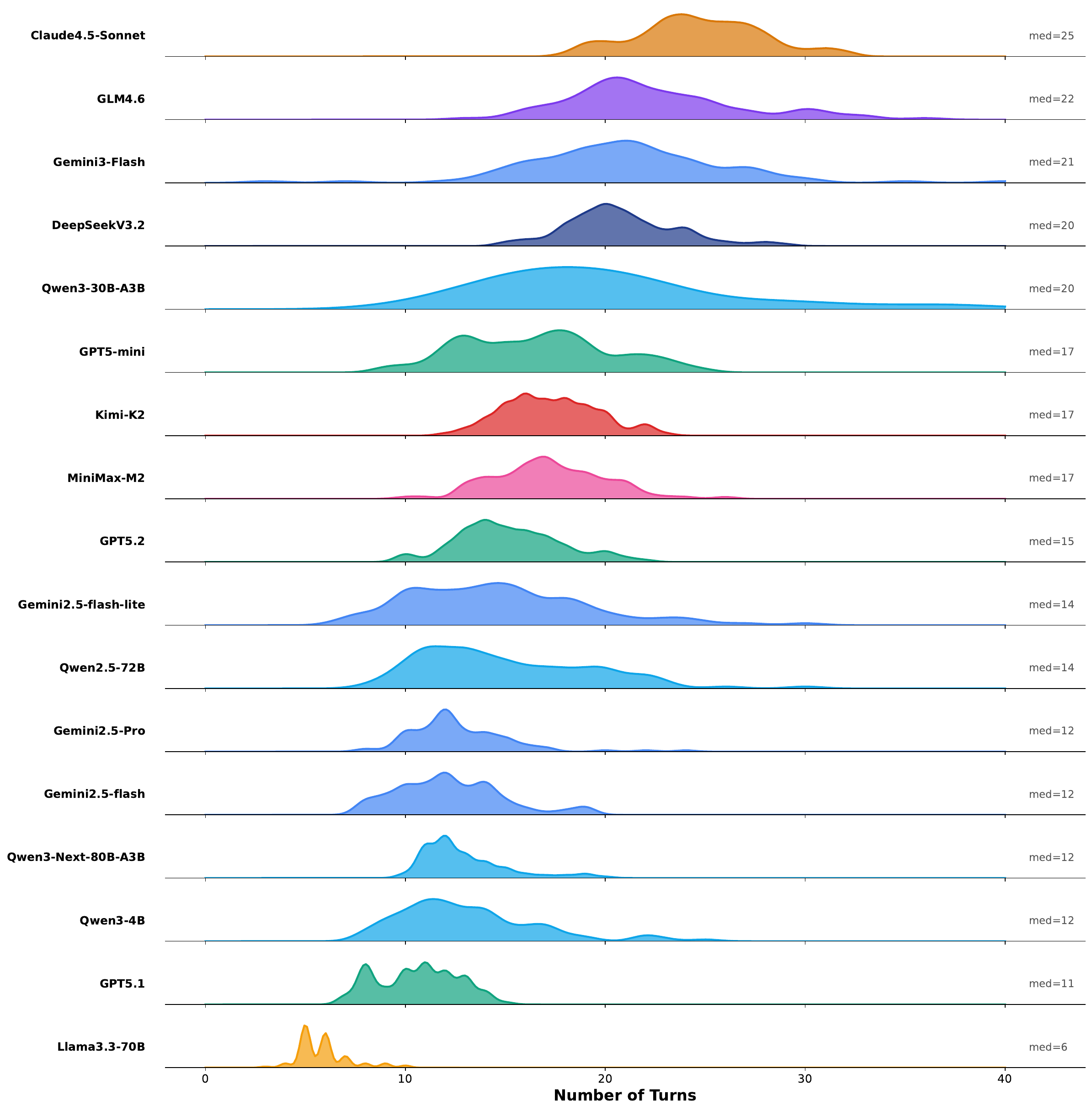}
    \caption{Interaction turn distribution on GLOBEM scenario.}
    \label{fig:turn_distribution_globem}
\end{figure*}

\begin{figure*}[htbp]
    \centering
    \includegraphics[width=0.9\linewidth]{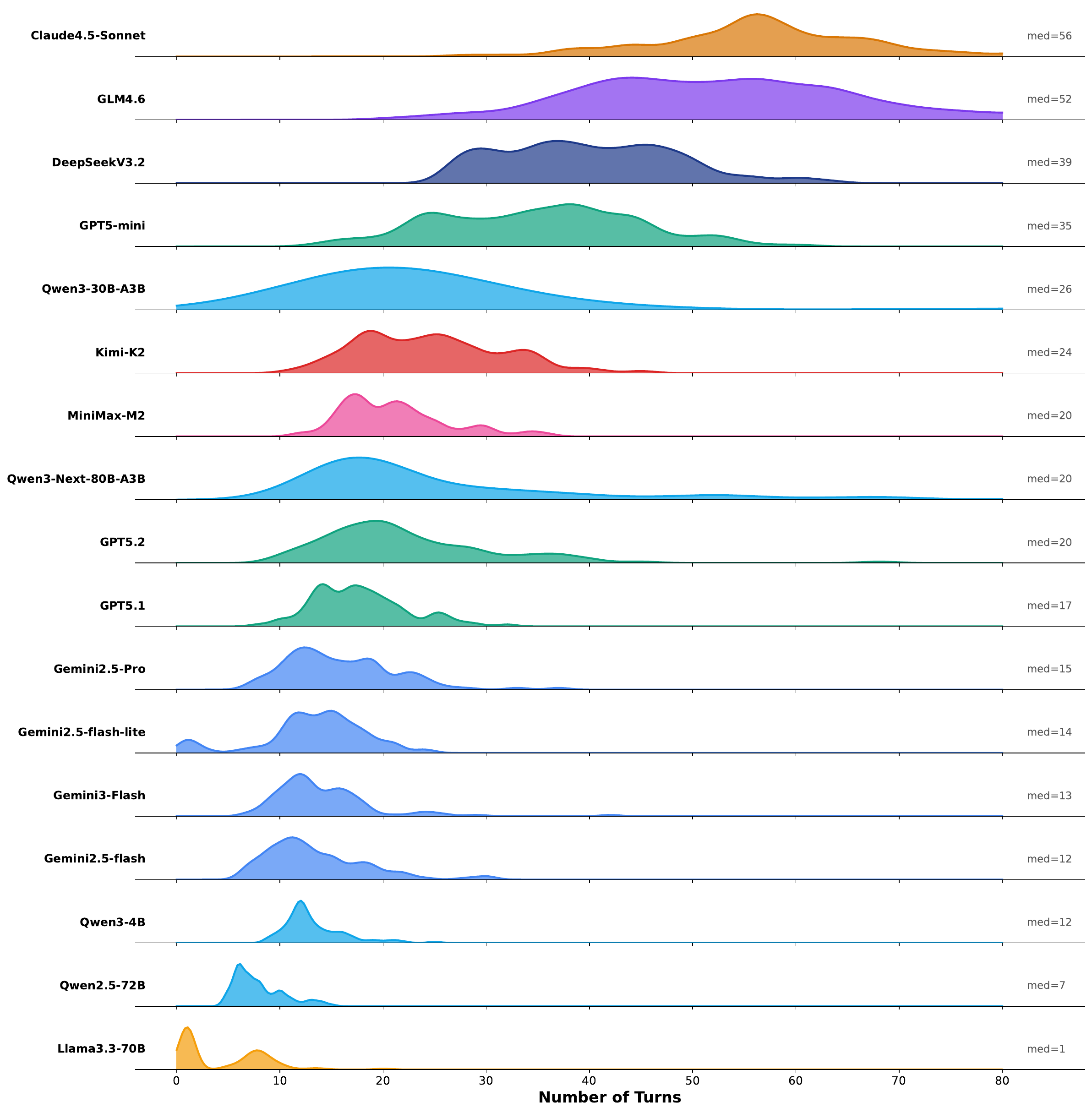}
    \caption{Interaction turn distribution on 10-K scenario.}
    \label{fig:turn_distribution_10k}
\end{figure*}

Figure~\ref{fig:turn_distribution_mimic}, \ref{fig:turn_distribution_globem} and \ref{fig:turn_distribution_10k} show the distributions of interaction rounds across models and scenarios. Almost all distributions are approximately normal, except for a small number of runs in which exploration failed to terminate, and the models repeatedly entered debugging loops. This behaviour was observed for a few models, such as Qwen3-30B-A3B and GLM4.6, and these runs were forcibly stopped at 100 rounds. We ignore these abnormal data points when drawing figures.

Stronger models tend to explore for more rounds without external prompting. Knowledge-intensive databases such as 10-K and MIMIC induce more interaction rounds than signal-based datasets such as GLOBEM, and the resulting distributions are also more uniform.

\section{Valid Insight Ratio}\label{appendix:valid_insight_ratio}
When generating message-wise insights, the LLM is prompted to decide autonomously whether the outcome of the current interaction is meaningful, as shown in Figure~\ref{fig:insight_prompt}. If it is not, the model outputs ``NO INSIGHT''. An interaction is treated as uninformative and marked as ``NO INSIGHT'' only when the content is:
\begin{enumerate}
    \item insufficient to support any insight
    \item failed function call
    \item when the interaction only invokes descriptive tools such as \verb|list_files|, \verb|describe_table|, \verb|get_database_info|, or \verb|get_field_description|
\end{enumerate}

The proportion of meaningful insights is then computed over all generated message-wise insights. In practice, at least two to three interactions are expected to involve descriptive tool calls and therefore produce no insight, which implies that the effective insight ratio cannot reach 100\%. 

\begin{figure*}[htbp]
    \centering
    \includegraphics[width=0.99\linewidth]{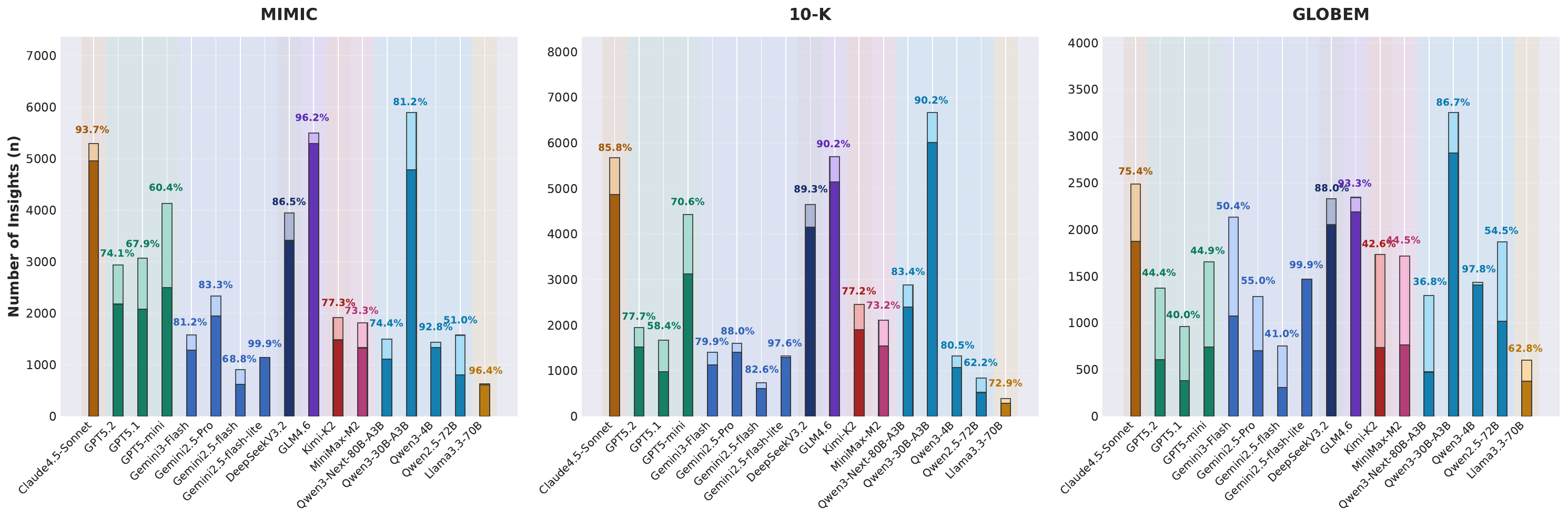}
    \caption{Valid insight ratio distribution.}
    \label{fig:valid_insight_ratio}
\end{figure*}

As shown in Figure~\ref{fig:valid_insight_ratio}, the height of each bar indicates the total number of message-wise insights generated, while the solid segment represents the number of valid insights. Stronger LLMs, including Claude 4.5 Sonnet, DeepSeek, and GLM, sustain a high effective insight ratio while generating a large volume of insights, which denotes that these models maintain high information density during the exploration.

\section{Tool Execution Time Distribution}\label{appendix:tool_exec_time}
\begin{figure*}[htbp]
    \centering
    \includegraphics[width=0.9\linewidth]{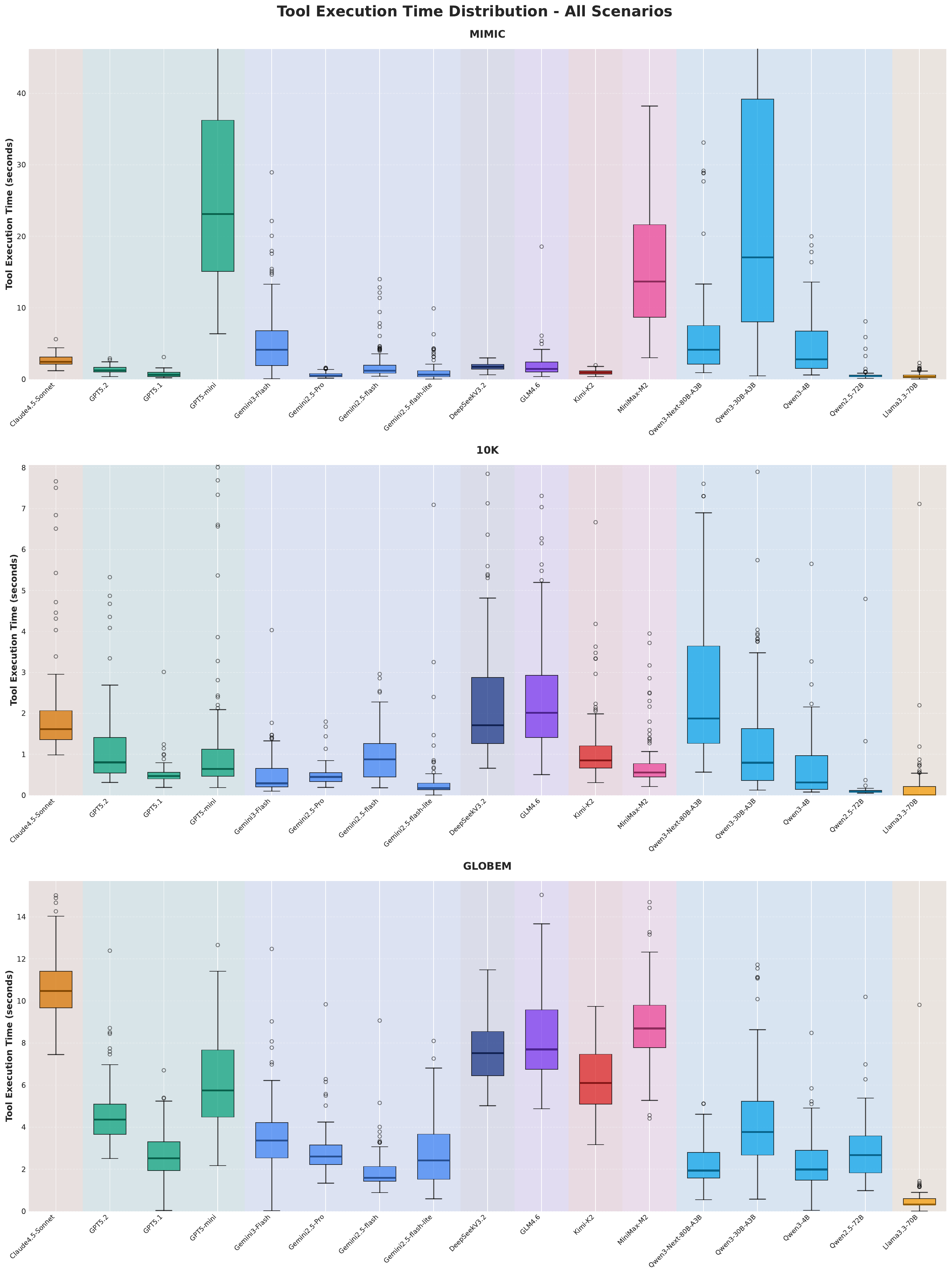}
    \caption{Average tool execution time per trajectory.}
    \label{fig:tool_exec_time}
\end{figure*}

As shown in Figure~\ref{fig:tool_exec_time}, the tool execution times of different models vary across scenarios. The duration of tool calls partially reflects the complexity of tool usage by each LLM. We report only the tool execution time, as the total runtime, including LLM generation, is affected by multiple factors such as API service stability and local GPU performance. Presenting the tool execution time allows for a fair comparison across different LLMs.

In the MIMIC scenario, GPT-5-mini and Qwen3-30B-A3B exhibit significantly longer tool execution times than other models, yet their overall performance does not surpass them. This indicates that complex queries do not necessarily yield more meaningful insights. Efficient and precise database interaction will be a crucial metric for evaluating a model’s data insight capabilities, encompassing high query efficiency (useful information returned per unit execution time), extraction of more information in fewer interaction rounds, and derivation of more meaningful insights from the same data.

\section{Hallucination-Accuracy Correlation}\label{appendix:hallucination_accuracy_correlation}
\begin{figure*}[htbp]
    \centering
    \includegraphics[width=0.99\linewidth]{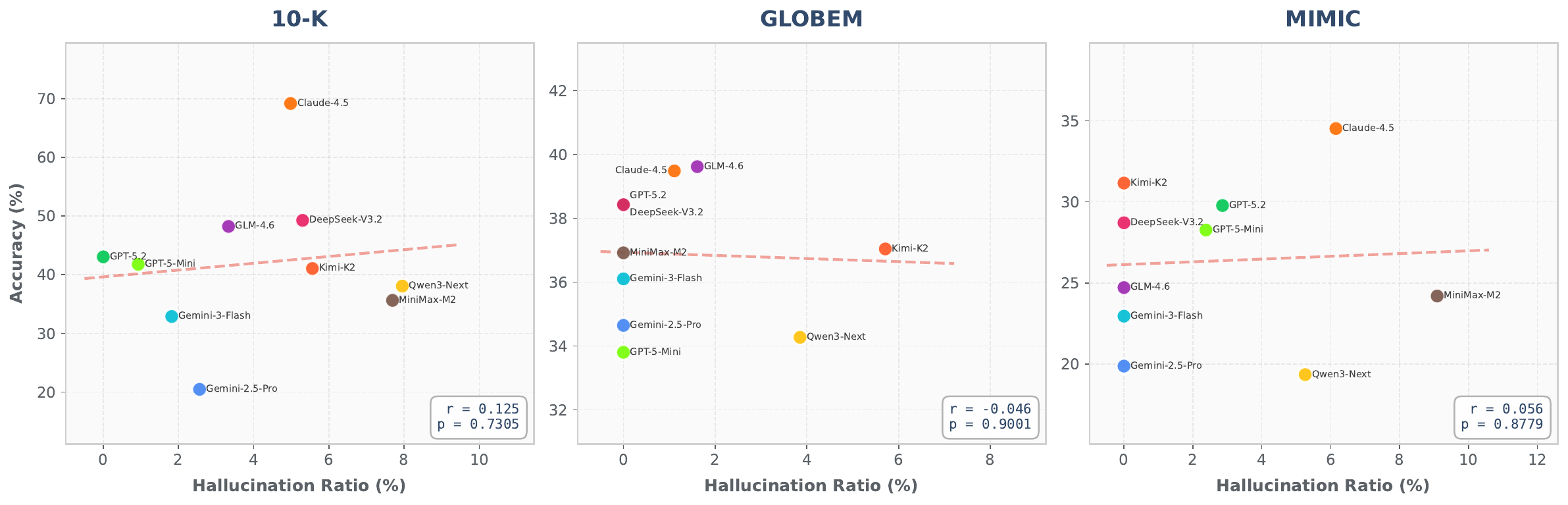}
    \caption{Hallucination-Accuracy Correlation.}
    \label{fig:hallucination_accuracy_correlation}
\end{figure*}

Figure~\ref{fig:hallucination_accuracy_correlation} shows that in the 10K and MIMIC settings, hallucination rates show extremely weak and non-significant positive correlations with accuracy, with correlation coefficients of 0.125 and 0.056, and $p$-values of 0.8779 and 0.7305, respectively. In the GLOBEM setting, the hallucination rate exhibits an extremely weak and non-significant negative correlation with accuracy, with a correlation coefficient of -0.046 and a $p$-value of 0.9001. Overall, the hallucination rate is essentially uncorrelated with final accuracy. It indicates that there is a very low risk that data contamination causes models to disregard actual database interaction results and inflate performance metrics.

\section{Prompts}
\subsection{System Prompt}\label{appendix:system_prompt}

Figure~\ref{fig:system_prompt} shows the system prompt used in DDR-Bench. It mainly enforces a ReAct-style multi-turn interaction and specifies strict requirements on the output format. Some constraints are repeatedly emphasised in uppercase, as smaller models otherwise struggle to follow the instructions and conduct reasonable multi-step agentic exploration. For larger models, the system prompt could be further simplified; however, for a fair comparison, we apply the same system prompt to all models.

\begin{figure*}[htbp]
\begin{tcolorbox}[title={System Prompt in DDR-Bench}]
You are an autonomous data analysis agent. Your task is to keep exploring and analyzing data for the given task.

\

IMPORTANT INSTRUCTIONS:

1. always respond in a ReAct style: return what you're thinking and planning to do, and then call the appropriate tool. RETURN BOTH THE TEXT CONTENT AND THE TOOL CALL.

2. you can only call one tool every turn.

3. your reasoning should contain insights derived from last turn's tool call results. BUT DO NOT INCLUDE ANY INSIGHTS OR REASONING IN THE TOOL CALLS. TOOLS ARE ONLY FOR DATA EXPLORE.

4. you should try your best to use the tools to get more information. keep exploring, build more and more complex params as turns go on and you will discover more in the data.

5. first use the tools to check what data is available to you.

\

TASK COMPLETION:
When you can not gather more information, send a message that starts with "FINISH:" followed by your all insights collected from the whole dialogue and tool calls. 

Only use "FINISH:" when you are absolutely certain that no more information can be gathered. 

Carefully use "FINISH:" in your message since it will immediately end the session. Think twice before using it.

\

YOUR TASK: \verb|{task}|

\

Analyze the task and use the available tools to accomplish it step by step.

\

ALWAYS RETURN in the format of text content with a tool call, no just return the tool call.
\end{tcolorbox}
\caption{System prompt in DDR-Bench. The system prompt give a detailed guideline to ensure that the LLMs follow the ReAct style for exploration, but it does not mention any specific question, task or goals. The task placeholder in this prompt denotes the simple start prompt part, like ``start analysis $\{entity\_id\}$''.}
\label{fig:system_prompt}
\end{figure*}

\subsection{Message-wise Insight Generation Prompt}\label{appendix:I_m_prompt}

Figure~\ref{fig:insight_prompt} shows the prompt used to generate message-wise insights. This prompt is independent from the agent trajectory: message-wise insight generation is not part of the agent’s ReAct trace. Instead, at each ReAct turn, we invoke this prompt to produce an insight and then continue the trajectory, while all calls are made to the same underlying model. The model only observes the agent’s ReAct content from the current turn and the corresponding environment execution results. Since we organise the multi-turn interaction between the agent and the environment in a dialogue format, the agent corresponds to the assistant and the environment to the user, which explains the use of the placeholders \texttt{assistant\_content} and \texttt{user\_content}.

\begin{figure*}[htbp]
\begin{tcolorbox}[title={Prompt for Generating Message-wise Insight $I_m$ in DDR-Bench}]
Based on the following tool execution, provide a brief insight about what was discovered or learned:

\

The reason and action to use the tool:

\verb|{assistant_content}|

\

Tool execution result:

\verb|{user_content}|

\

Provide a concise insight (1-3 sentences) about what this reveals:

1. It has to be related to the task: {task}.

2. If there is no insight or error in the tool execution, respond with 'NO INSIGHT'.

3. If it only use the data description tools (e.g. tools like \verb|list_files, describe_table|

\verb|, get_database_info, get_field_description|), respond with 'NO INSIGHT'.

4. The insight from data should answer the question raised in the reason to execute this tool. Focus on this point.

5. Keep all the data or statistics needed in your generated insight.
ONLY respond with the insight.

\end{tcolorbox}
\caption{Prompt for Generating Message-wise Insight $I_m$ in DDR-Bench.}
\label{fig:insight_prompt}
\end{figure*}

\subsection{Novelty Pairwise Judge Prompts}\label{appendix:novelty_judge_prompts}

Figure~\ref{fig:novelty_judge_mimic}, \ref{fig:novelty_judge_10k}, and \ref{fig:novelty_judge_globem} provide the prompt for the pairwise novelty judge in the scenario of MIMIC, 10-K and GLOBEM, respectively.

\begin{figure*}[htbp]
\begin{tcolorbox}[title={Novelty Pairwise Judge Prompt for the MIMIC Scenario}]
You are an expert clinical evaluator. You will compare two sets of novel clinical insights about the same patient.

\

Patient Context: These insights were generated during clinical data analysis but were NOT used to answer specific clinical questions correctly. They represent potentially valuable but unused observations.

\

Your task: Determine which set provides MORE clinically actionable and useful information for understanding the patient's condition.

\

Consider:

- Diagnostic value: Does it help identify or clarify diagnoses?

- Treatment implications: Does it inform treatment decisions?

- Prognostic relevance: Does it help predict outcomes?

- Clinical actionability: Can clinicians act on this information?

- Depth of insight: Does it reveal meaningful patterns or connections?

- Do not be biased by the length, number of insights, fluency, etc. Just focus on the usefulness of the insights.

\

Insights from Model A:

\{insights\_a\}

Insights from Model B:

\{insights\_b\}

\

Respond in EXACTLY this format (two lines):

Line 1: One sentence explaining your reasoning (max 100 words)

Line 2: Your decision - ONLY one of: MODEL\_A, MODEL\_B, or TIE

\

Example:
Model A provides more specific diagnostic details about the patient's cardiovascular condition and treatment implications.

MODEL\_A

\

Your response:

\end{tcolorbox}
\caption{Novelty Pairwise Judge Prompt for the MIMIC Scenario.}
\label{fig:novelty_judge_mimic}
\end{figure*}

\begin{figure*}[htbp]
\begin{tcolorbox}[title={Novelty Pairwise Judge Prompt for the 10-K Scenario}]
You are an expert financial analyst. You will compare two sets of novel financial insights about the same company.

\

Company Context: These insights were generated during 10-K analysis but were NOT used to answer specific financial questions correctly. They represent potentially valuable but unused observations.

\

Your task: Determine which set provides MORE valuable information for investment analysis and business understanding.

\

Consider:

- Investment value: Does it inform investment decisions?

- Business strategy implications: Does it reveal strategic directions or challenges?

- Financial health indicators: Does it highlight financial strengths or risks?

- Competitive positioning: Does it clarify market position or advantages?

- Depth of insight: Does it reveal meaningful patterns or trends?

- Do not be biased by the length, number of insights, fluency, etc. Just focus on the usefulness of the insights.

\

Insights from Model A:

\{insights\_a\}

Insights from Model B:

\{insights\_b\}

\

Respond in EXACTLY this format (two lines):

Line 1: One sentence explaining your reasoning (max 100 words)

Line 2: Your decision - ONLY one of: MODEL\_A, MODEL\_B, or TIE

\

Example:
Model A provides more specific financial metrics and strategic insights that are more actionable for investment decisions.

MODEL\_A

\

Your response:

\end{tcolorbox}
\caption{Novelty Pairwise Judge Prompt for the 10-K Scenario.}
\label{fig:novelty_judge_10k}
\end{figure*}

\begin{figure*}[htbp]
\begin{tcolorbox}[title={Novelty Pairwise Judge Prompt for the GLOBEM Scenario}]
You are an expert behavioral psychologist and mental health researcher. You will compare two sets of novel behavioral insights about the same user.

\

User Context: These insights were generated during passive sensing data analysis but were NOT used to answer specific mental health questions correctly. They represent potentially valuable but unused observations.

\

Your task: Determine which set provides MORE useful understanding of their mental health patterns and behavioral trends.

\

Consider:

- Behavioral indicators: Does it identify meaningful behavior patterns?

- Mental health relevance: Does it relate to psychological well-being?

- Longitudinal trends: Does it reveal changes over time?

- Actionability: Can this inform interventions or support?

- Depth of insight: Does it reveal meaningful connections in the data?

- Do not be biased by the length, number of insights, fluency, etc. Just focus on the usefulness of the insights.

\

Insights from Model A:

\{insights\_a\}

Insights from Model B:

\{insights\_b\}

\

Respond in EXACTLY this format (two lines):

Line 1: One sentence explaining your reasoning (max 100 words)

Line 2: Your decision - ONLY one of: MODEL\_A, MODEL\_B, or TIE

\

Example:
Model A identifies more specific behavioural patterns related to mental health that could inform targeted interventions.

MODEL\_A

\

Your response:

\end{tcolorbox}
\caption{Novelty Pairwise Judge Prompt for the GLOBEM Scenario.}
\label{fig:novelty_judge_globem}
\end{figure*}

\subsection{LLM-as-a-Checker Prompts}\label{appendix:acc_judge_prompts}
Across the three scenarios, the LLM is prompted to act as a checker according to the same principle, namely, to assess whether the provided insights are sufficient to support answering the questions specified in the checklist items, including
\begin{enumerate}
    \item Determine if the messages can provide evidence to support the answer
    \item Identify which specific message(s) by their index numbers \verb|[Message X]| support or contradict the answer
    \item Extract the evidence text from the relevant message(s)
    \item Classify the context quality into one of three categories: \verb|CORRECT_INFO|: Messages contain information that serves as evidence or support for the answer; \verb|INCORRECT_INFO|: Messages contain information that contradicts the answer; \verb|INSUFFICIENT_INFO|: Messages lack sufficient information to answer the question.
\end{enumerate}
The proportion of \verb|CORRECT_INFO| is calculated as the final accuracy. In addition, the GLOBEM scenario uses a closed-form checklist and therefore does not require the LLM for checking. Instead, the LLM directly answers the questions based on the insights, and accuracy is computed by exact match against the ground truth. The reason that GLOBEM and 10 K do not adopt a sampling-based checking procedure is that answers to open-ended questions are difficult to normalise.

Furthermore, the prompt includes scenario-specific evaluation criteria as well as JSON return fields to facilitate downstream data processing. Details can be found in the project code at \verb|evaluate/prompts.py|.

\section{Trajectory Sample}\label{appendix:traj_sample}
\begin{figure*}[htbp]
\begin{tcolorbox}[title={Trajectory Sample on MIMIC}]
\footnotesize

\includegraphics[width=12pt]{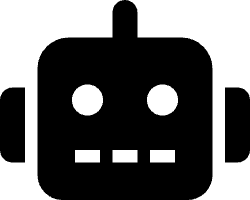} Agent: I'll help you analyze patient xxxxxx. Let me start by exploring what data is available in the database. 

execute:  \verb|{"tool": "get_database_info", "arguments": {}}|

\

\includegraphics[width=12pt]{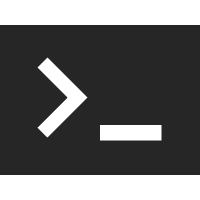} Environment: \verb|{'database_path': 'dataset/mimic.db', 'table_count': 27, 'tables':['hosp_a |

\verb|dmissions', 'hosp_d_hcpcs'......], 'tables_info': [{'name': 'hosp_admissions', |

\verb|'comment': "Hospital admissions for each patient. Defines......|

\

\includegraphics[width=12pt]{figs/agent.png} Agent: Good! I can see this is a MIMIC database with 27 tables containing patient medical data. Now let me search for patient xxxxxx starting with the patient demographics table. 

execute: \verb|{"tool": "execute_query", "arguments": {"query": "SELECT * FROM hosp_patients |

\verb| WHERE subject_id = xxxxxx"}|

\

\includegraphics[width=12pt]{figs/env.png} Environment: \verb|{'cols': ['subject_id', 'gender', 'anchor_age', 'anchor_year', 'anchor |

\verb|_year_group', 'dod'], 'data': [['xxxxxx', 'M', '74', '2182', '2008 - 2010', |

\verb| '2185-06-15']], 'count': 1}|

\

\includegraphics[width=12pt]{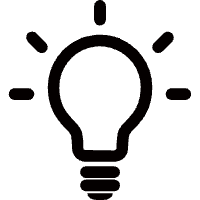} Agent Message-wise Insight: Patient xxxxxx is a 74-year-old male with an anchor year of 2182 (corresponding to the real-world period 2008-2010) who died on June 15, 2185, indicating he survived approximately 3 years beyond the anchor year recorded in this database.

\

\textit{...(46 interactions)}

\

\includegraphics[width=12pt]{figs/agent.png} Agent: No prior TIA documented. Let me check CHA2DS2-VASc risk factors systematically for stroke risk with AFib.

execute: \verb|{"tool": "execute_query", "arguments": {"query": "SELECT \n  'Has CHF' as  |

\verb|risk_factor, COUNT(DISTINCT d.hadm_id) as admissions_with_dx\nFROM hosp_diagnoses_icd  |

\verb|d\nWHERE d.subject_id = xxxxxx AND d.icd_code LIKE '428%'\nUNION ALL\nSELECT 'Has HTN',  |

\verb|COUNT(DISTINCT d.hadm_id)\nFROM hosp_diagnoses_icd d\nWHERE d.subject_id =......|

\

\includegraphics[width=12pt]{figs/env.png} Environment: \verb|{'cols': ['risk_factor', 'admissions_with_dx'], 'data': [['Has CHF', '15'],  |

\verb|['Has HTN', '18'], ['Has DM', '17'], ['Has Vascular Dz', '9'], ['Has AFib', '16']],  |

\verb|'count': 5}|

\

\includegraphics[width=12pt]{figs/idea.png} Agent Message-wise Insight: This patient (ID: xxxxxx) has documented atrial fibrillation and multiple CHA2DS2-VASc risk factors present across their 18 admissions, including hypertension (18 admissions), diabetes (17 admissions), CHF (15 admissions), and vascular disease (9 admissions), indicating a high stroke risk profile that would warrant anticoagulation therapy consideration.

\

\textit{...(12 interactions)}

\

\includegraphics[width=12pt]{figs/idea.png} Agent Trajectory-wise Insight: Excellent! I now have a comprehensive understanding of Patient xxxxxx. Let me compile all the insights.
FINISH: 

\begin{verbatim}
## COMPREHENSIVE ANALYSIS OF PATIENT xxxxxx

### DEMOGRAPHIC PROFILE
- **Patient ID**: xxxxxx
- **Gender**: Male
- **Age at anchor**: 74 years (anchor year 2182)
- **Age at death**: 77 years
- **Date of death**: June 15, 2185
......

This patient exemplifies the challenges of managing elderly patients with multiple,

severe chronic conditions and the high mortality risk associated with acute stroke 

in the setting of advanced multi-organ disease.
\end{verbatim}

\end{tcolorbox}
\caption{Trajectory Sample on MIMIC}
\label{fig:traj_sample_mimic}
\end{figure*}

\begin{figure*}[htbp]
\begin{tcolorbox}[title={Trajectory Sample on GLOBEM}]
\footnotesize

\includegraphics[width=12pt]{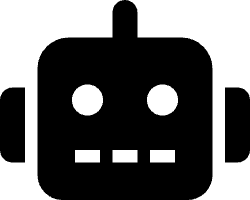} Agent: I'll help you analyze how user behavior patterns change over time for user xxxxxx... 

execute:  \verb|{"tool": "list_files", "arguments": {}}|

\

\includegraphics[width=12pt]{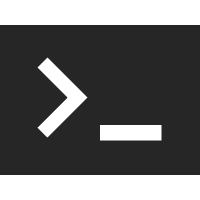} Environment: \verb|{'files': [{'name': 'activity_allday_raw.csv', 'path': ' |

\verb|activity_allday_raw.csv', 'type': 'file', 'size': 2067543, 'modified':  |

\verb|'2025-11-21T20:08:33', 'description': 'Daily activity metrics including movement and  |

\verb|exercise patterns.'}......], 'count': 6, 'pattern': None, 'recursive': True}|

\

\textit{...(19 interactions)}

\

\includegraphics[width=12pt]{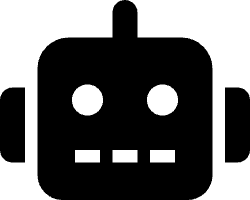} Agent: Excellent! I can see a clear behavioral shift starting April 30th. Let me now examine circadian rhythm patterns and correlate multiple behavioral dimensions.

execute: 
\begin{lstlisting}
import pandas as pd
import numpy as np
user_pid = 'xxxxxx'
# Merge multiple data sources for correlation analysis
phone_df = pd.read_csv('phone_usage_allday_raw.csv')
loc_df = pd.read_csv('location_allday_raw.csv')
conn_df = pd.read_csv('connectivity_allday_raw.csv')
# ...(66 lines)
for label, (metric, before_val, after_val) in metrics.items():
    if pd.notna(before_val) and pd.notna(after_val) and before_val != 0:
        change = ((after_val - before_val) / before_val * 100)
        print(f\"{label:<30}:{change:>+7.1f}%\")
\end{lstlisting}

\

\includegraphics[width=12pt]{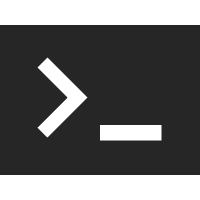} Environment: \verb|Critical Behavioral Shift: Day 28 (April 30, 2018) BEFORE SHIFT |

\verb| (Days 1-27): |

\verb| Phone unlocks:       34.3 per day |

\verb| Screen time:         87.7 minutes |

\verb| ...... |

\verb| Distance traveled             :  +211.2\% |

\verb| Location transitions          :   +66.4\% |

\verb| Social proximity scans        :   +51.6\% |

\

\includegraphics[width=12pt]{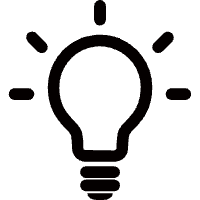} Agent Message-wise Insight: A dramatic behavioral shift occurred on Day 28 (April 30, 2018) for user xxxxxx, indicating a potential psychological status change. Phone unlocks increased 197.7\% (34.3→102.1/day), screen time rose 134\% (87.7→205.3 min/day), first phone use shifted 77.4\% earlier (434→98 min after midnight), distance traveled surged 211.2\% ......

\

\textit{...(6 interactions)}

\

\includegraphics[width=12pt]{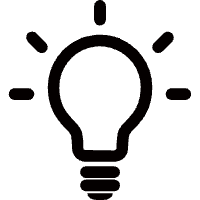} Agent Trajectory-wise Insight: Excellent! I have completed a comprehensive analysis. I have gathered extensive information about the user's behavioral patterns and their changes over time. Let me now provide a final comprehensive summary.
FINISH: 

\begin{verbatim}
# COMPREHENSIVE BEHAVIORAL ANALYSIS: USER xxxxxx

## Overview
I analyzed 92 days of multimodal behavioral data (April 3 - July 3, 2018) for

user xxxxxx, examining patterns across 6 behavioral domains......User xxxxxx exhibited 

a **statistically significant, clinically meaningful, multi-dimensional behavioral

transformation** starting precisely on Day 28 (April 30, 2018). The pattern...


\end{verbatim}

\end{tcolorbox}
\caption{Trajectory Sample on GLOBEM}
\label{fig:traj_sample_globem}
\end{figure*}

\begin{figure*}[htbp]
\begin{tcolorbox}[title={Trajectory Sample on 10-K}]
\footnotesize

\includegraphics[width=12pt]{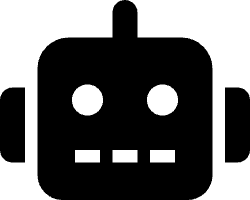} Agent: I'll analyze the structured financial data from Phillips 66 (CIK: 1534701). Let me start by exploring what data is available in the database.

execute:  \verb|{'database_path': 'dataset/10-k.db', 'table_count': 4, 'tables': ['companies',  |

\verb|'company_addresses', 'company_tickers', 'financial_facts'..., 'filtered_out_count': 5}}|

\

\textit{...(8 interactions)}

\

\includegraphics[width=12pt]{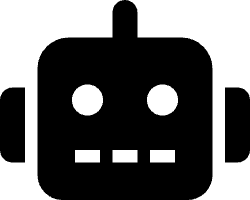} Agent: Perfect! Now I can see the pattern. For fiscal year 2024, the end\_date is 2024-12-31, and the revenue is \textdollar 143.153 billion. The highest value shown (like \textdollar 169.99B for end\_date 2022-12-31) are historical comparatives. Let me now get a clean view of annual revenues and net income by matching fiscal\_year to end\_date year.

execute: \verb|{"tool": "execute_query", "arguments": {"query": "SELECT fiscal_year, fact_name,  |

\verb|fact_value, end_date FROM financial_facts WHERE cik = '1534701' AND fiscal_period =  |

\verb|'FY' AND fact_name IN ('RevenueFromContractWithCustomerExcludingAssessedTax',  |

\verb|'NetIncomeLoss') AND CAST(strftime('%Y', end_date) AS INTEGER) = fiscal_year ORDER BY  |

\verb|fiscal_year DESC, fact_name", "limit": 50}|

\

\includegraphics[width=12pt]{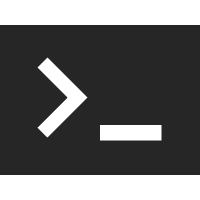} Environment: \verb|{'cols': ['fiscal_year', 'fact_name', 'fact_value', 'end_date'], 'data':  |

\verb|[['2024', 'NetIncomeLoss', '2117000000.0', '2024-12-31'], ['2024',  |

\verb|'RevenueFromContractWithCustomerExcludingAssessedTax', '143153000000.0', '2024-12-31'],  |

\verb|['2023', 'NetIncomeLoss', '7015000000.0', '2023-12-31'], ['2023',  |

\verb|'RevenueFromContractWithCustomerExcludingAssessedTax', '147399000000.0', '2023-12-31'] |

\verb|...], 'count': 50}|

\

\includegraphics[width=12pt]{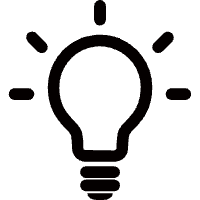} Agent Message-wise Insight: Phillips 66 experienced significant revenue volatility over the past five years, with revenues peaking at \textdollar170.0 billion in 2022 before declining 15.8\% to \textdollar143.2 billion in 2024. More concerning is the sharp decline...

\

\textit{...(70 interactions)}

\

\includegraphics[width=12pt]{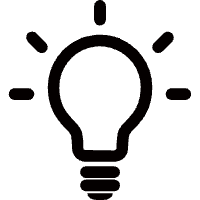} Agent Trajectory-wise Insight: Perfect! Now I have a complete picture of Phillips 66's financial data. I have gathered extensive information covering:
- Income statement data (2012-2024)
- Balance sheet data (2012-2024)
- Cash flow data (2012-2024)
- Quarterly trends (2023-2025 Q2)
- Key metrics like EPS, dividends, share repurchases, debt, working capital, etc.
Now I can provide comprehensive insights based on this data.
FINISH: Based on my comprehensive analysis of Phillips 66's (CIK: 1534701) structured financial data from 2012-2025 (through Q2 2025), here are the material quantitative patterns, trends, and relationships that investors should consider:

\begin{verbatim}
# REVENUE AND PROFITABILITY TRENDS

**Revenue Volatility Pattern (2018-2024):**
- Peak: $169.99B in 2022 (+52.5% from 2021's $111.48B)
- Decline: $147.40B in 2023 (-13.3% YoY)
- Further decline: $143.15B in 2024 (-2.9% YoY)
- This reveals significant sensitivity to commodity price cycles, with 2022 benefiting 
from energy price spikes post-pandemic

**Net Income Cyclicality:**
- 2020: -$539M (loss year during pandemic)
- 2021: $1.32B (recovery but weak)

......

The data reveals Phillips 66 is in a challenged operating environment with compressed
margins, declining profitability, and aggressive capital returns that may not be
sustainable at current earnings levels. The company is drawing on balance sheet
strength built during 2022's exceptional year, but with declining cash, rising
leverage, and deteriorating coverage ratios, financial flexibility is eroding.
Investors should assess whether the current 27.8x P/E ratio appropriately reflects both
cyclical risks and the company's strategic positioning in a potentially structurally
challenged refining industry.
\end{verbatim}

\end{tcolorbox}
\caption{Trajectory Sample on 10-K}
\label{fig:traj_sample_10k}
\end{figure*}

Figure~\ref{fig:traj_sample_mimic}, \ref{fig:traj_sample_10k}, and \ref{fig:traj_sample_globem} provide a trajectory sample in the scenario of MIMIC, 10-K and GLOBEM, respectively. The trajectory includes the reasoning, action, observation (execution results from the database environment) and two kinds of insights, message-wise insight $I_m$ and trajectory-wise insight $I_t$. We anonymise the task entity id in the MIMIC and GLOBEM.

\end{document}